\documentclass[sn-mathphys]{sn-jnl}

\jyear{2021}%

\theoremstyle{thmstyleone}

\theoremstyle{thmstyletwo}

\theoremstyle{thmstylethree}%

\usepackage{makecell}
\usepackage{graphicx}
\usepackage{float} 
\usepackage{subfigure}

\begin{document}

\title[Hard Example Guided Hashing for Image Retrieval]{Hard Example Guided Hashing for Image Retrieval}


\author[1]{\fnm{Hai} \sur{Su}}\email{suhai@m.scnu.edu.cn}

\author[1]{\fnm{Meiyin} \sur{Han}}\email{hanmeiyin@m.scnu.edu.cn}

\author[2]{\fnm{Junle} \sur{Liang}}\email{juliang@syr.edu}

\author*[1]{\fnm{Jun} \sur{Liang}}\email{liangjun@m.scnu.edu.cn}

\author[1]{\fnm{Songsen} \sur{Yu}}\email{yss8109@163.com}

\affil[1]{\orgdiv{School of Software}, \orgname{South China Normal University}, \orgaddress{\street{Taoyuan East Road}, \city{Foshan}, \postcode{528225}, \state{Guangdong}, \country{China}}}

\affil[2]{\orgdiv{College of Engineering and Computer Science}, \orgname{Syracuse University}, \orgaddress{\street{South Crouse Ave}, \city{Syracuse}, \postcode{13255}, \state{New York}, \country{United States}}}


\abstract{Compared with the traditional hashing methods, deep hashing methods generate hash codes with rich semantic information and greatly improves the performances in the image retrieval field. However, it is unsatisfied for current deep hashing methods to predict the similarity of hard examples. It exists two main factors affecting the ability of learning hard examples, which are weak key features extraction and the shortage of hard examples. In this paper, we give a novel end-to-end model to extract the key feature from hard examples and obtain hash code with the accurate semantic information. In addition, we redesign a hard pair-wise loss function to assess the hard degree and update penalty weights of examples. It effectively alleviates the shortage problem in hard examples. Experimental results on CIFAR-10 and NUS-WIDE demonstrate that our model outperformances the mainstream hashing-based image retrieval methods.}

\keywords{Image Retrieval $\cdot$ Deep Hashing $\cdot$ Hard Examples}

\maketitle

\section{Introduction}\label{sec1}

In recent years, retrieval technology has become an important tool for retrieving the required objects from large-scale image databases. As a branch of image retrieval, Approximate Nearest Neighbor (ANN) method has caught the researcher’s attention with high retrieval precision and low computation cost. Hashing retrieval, a representative method of ANN, behaves the high speed of retrieval because of reduced cost of storage. It transforms high dimension features of an image into a short hash code and utilizes hamming distance of hash code to represent the similarity between images. The closer the hamming distance between two hash codes, the more similar the images are. Otherwise, the less similar they are.

The order of hamming distance between hash codes generated by well-designed hash function, should be as much consistent as possible with the order of image sorting by images similarity. In traditional approaches, researchers design hand-crafted hash function, and encode hash code by projecting example data on a new feature space \cite{bib1,bib2}. Alternatively, the traditional hashing methods divide the original space of data according to the number of bits in the hash code and determine the value of each bit by the position of the examples relative to each hyperplane \cite{bib3,bib4}. Those methods are too simple to handle with tremendous and complex data. Deep hashing, proposed to be a remedy of limitation of tradition method, integrates more semantic information to generate hash codes by applying CNN \cite{bib5,bib6,bib7,bib8,bib9}. The significant performance of deep hashing progress indicates a great potential for image retrieval. 

\begin{figure}[H]
    \centering
    \includegraphics[width = \textwidth]{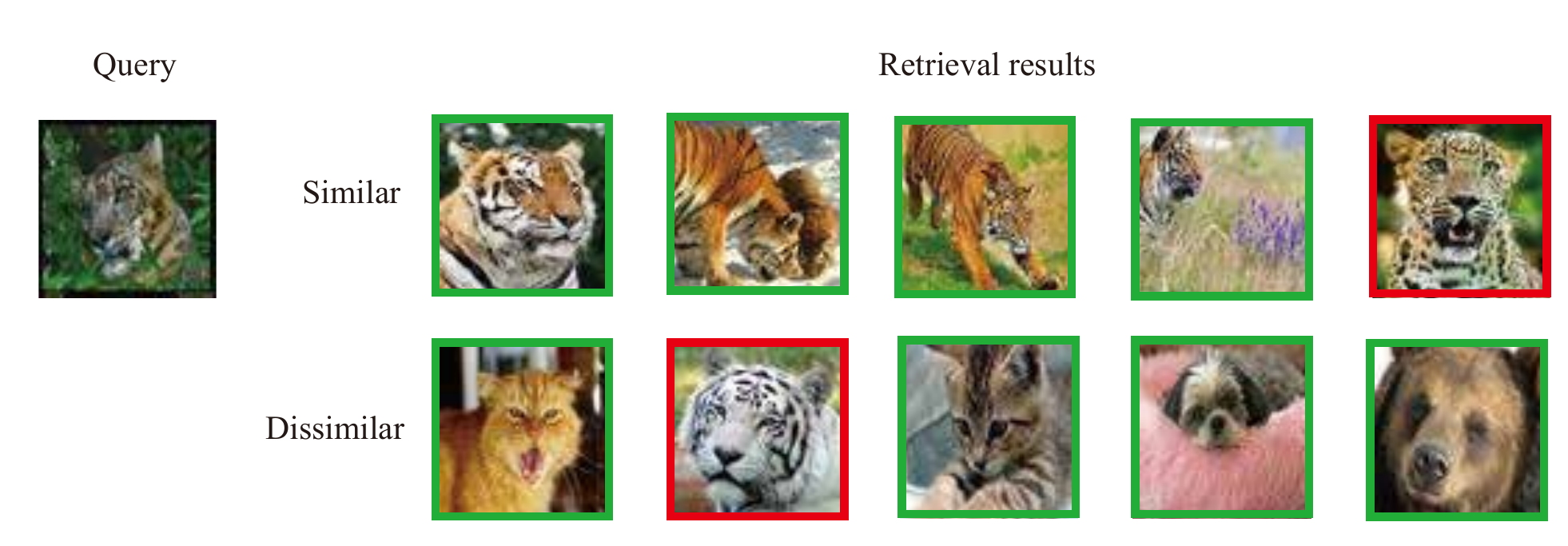}
    \caption{The example of the hard example from NUS-WIDE. The left is the image to be retrieved, and the right is the search result. Similar images calculated by deep hashing are in the first column and dissimilar images are in the second column. The green boxes indicate that the similarity calculated by deep hashing is correct, and the red box represents that the results are wrong.}\label{fig.1}
\end{figure}

However, we find that the hard example is an issue within current deep hashing methods. Current methods cannot judge the similarity of hard examples well \cite{bib10}. Hard examples are hard to identify correctly and lead to unexpected result. They generally are categorized as hard similar examples and hard dissimilar examples. Hard similar examples mean similar examples that are recognized as dissimilar by methods because of their different fine-grained surface. Conversely, hard dissimilar examples mean that are dissimilar semantically but would be regarded as similar because of the similarity of color, outline, or other coarse-grained outlooks. Fig. \ref{fig.1} shows the hard example from NUS-WIDE. As an example of the hard dissimilar example, Leopards and Bengal tigers are considered similar because of their similarity in coarse-grained features such as color and posture. Contrarily, the white Bengal tiger is mistaken for dissimilar because of the different color when the query image is a yellow Bengal tiger. The hard example is a special case, and its hard lies in the confusion caused by coarse-grained features. The existence of hard examples limits further improvement of hash retrieval performance.

The main reason for the misjudgment of similarity of hard examples is that hashing methods do not find distinguished high-level features. Such high-level features are often in hard examples. At present, the optimization of hashing methods is dominated by easy examples, and most of the features learned are low-level features. Therefore, we try to guide the network to learn hard examples to improve the retrieval accuracy.

Because of the interference of redundant information and the shortage of hard examples, hard examples are difficult to be learned. A lot of unnecessary information changes the dominance of key features and prevents learning key features in hard examples. In addition, it is small the amount of hard examples in most datasets. The information of hard examples in backpropagation is often ignored. Therefore, there are two main factors that affect the ability of the network to learn hard examples: weak key features extraction and the shortage of hard examples.

Based on those points, firstly, key feature is important to improve the ability of the network to learn hard examples. As we know, most of the key features are in the target area. However, the current network used in image retrieval treats all channels and regions indiscriminately, resulting in the final extracted features mix up with unnecessary information. Redundant information is deleterious for the retrieval task, because it will weaken or even bury the key features, which affects the precision. Therefore, we hope that the network learns the key features of hard examples, which effectively avoids generating hash codes with incorrect semantics. 

Secondly, the shortage of hard examples is a problem we have to face. It causes two risks. In the first place, the tiny quantity of certain examples reduces the detection ability of the model for these examples. The model moves the decision boundary towards them and result in over-fitting phenomenon. In addition, due to the shortage, hard examples are not dominant in the back propagation, or even are oversighted. The perfect solution is to collect more hard examples, but they are often difficult to obtain in real environment. Therefore, the problem is solved by other ways in this paper.

Reference \cite{bib9} and \cite{bib10}, they discuss the problem of hard examples first time. Based on the idea of hard mining, the authors resolve the shortage by increasing penalties for hard examples. Hard examples are selected by manually setting a threshold value of hamming distance. Hamming distance is calculated by discretized hash code. The calculation loses a lot of information. Thus, this method of selection is not conducive. After selecting the hard examples, they simply increase the punishment weight of the hard examples, without considering the influence of key feature on the learning of hard examples. Based on this, we propose a novel deep hashing method, called Hard Example Guided Hashing for Image Retrieval (HEGH), which improves the ability of the network to learn hard examples while preserving the similarity of the easy example. HEGH solves the problem from two aspects, weak key features extraction and the shortage of hard examples. The main contributions are as follows:

\begin{itemize}
    \item We design a new network, CNNs with CBAM, to locate the target, in order to further extract key feature from hard example. Our network performs well on focusing on target of hard example, so that the difficulty of the hard examples learning is reduced and the generated hash code has more accurate semantic information.
    
    \item For the selection of hard examples, we evaluate the hard degree of examples according to the classification ability of the network, which serves for the distribution of punishment weights of examples.
    
    \item A hard pair-wise loss function is proposed to alleviate the imbalance of example quantity. The loss function reduces the penalty weights of the easy examples, so that the optimization can be dominated by the hard example. 
\end{itemize}

The rest of the paper is organized as follows: Section \ref{sec2}  briefly discusses the related works. In Section \ref{sec3}, we describe our new method HEGH in detail. Experiment results and analyses are presented in Section \ref{sec4}. Section \ref{sec5} gives the conclusion of the paper.

\section{Related Work}\label{sec2}

The existing hash methods can be divided into unsupervised and supervised according to the existence of supervised information. Unsupervised hash retrieval algorithm takes an unlabeled dataset as input to train hash function and generates binary hash code \cite{bib11,bib12,bib13}. The unsupervised method quantifies the error loss indirectly and difficultly, which means that it exists semantic gap in the training process due to the lack of label information. Contrarily, deep supervised hash makes full use of label information and generates better hash codes with short hash codes and higher accuracy than unsupervised hash methods.

Most deep supervised hash methods focus on two main problems to achieve improvement: easing discrete optimization of hash codes and sustaining pair-similarity of data in hamming space. At present, certain research achievements have been made on the first problem \cite{bib14,bib15,bib16,bib17}, and some breakthrough has been made. DSDH \cite{bib17} uses a discrete cyclic coordinate descend (DCC) algorithm to solve the discrete optimization problem and keeps the discrete constraint in the whole optimization process. GreedyHash \cite{bib16} adopted a greedy algorithm to deal with the discrete hash optimization problem by replacing the quantization error with the hash layer of sign function. 

To maintain the semantic similarity of data in Hamming space, the various loss functions can be divided into pair-wise \cite{bib18,bib19,bib20,bib21}, triplet-wise \cite{bib5,bib8,bib11}. These loss functions concrete the image similarity order base on similarity information and ensure the original space and Hamming space as consistent as possible. The pair-wise loss function primarily constrains the distance between the similar example pairs within a fixed radius circle, using square loss and cross-entropy loss. Triplet-wise loss primarily constrains the difference of the distance between dissimilar example pair and similar example pair not less than a fixed value. It indicates obviously triplet-wise require much more preprocessed images. As a result, pair-wise loss function is a popular and efficient loss function. In the paper, we resign a novel loss function based on the pair-wise loss function.

The current retrieval methods rank effectively the similarity of easy examples, except hard examples. Easy examples are similar in appearance and belong to the same semantic class, which is understood on the basis of low-level features. But hard examples are difficult to be comprehended. Current methods rank hard similar example after hard dissimilar examples. Enhancing the learning of hard examples can help the methods study more high-level features, instead of judging semantics only by coarse-grain features. It can effectively optimize the order of the retrieval sequence.

Currently, problems of hard examples are found in image classification and target detection \cite{bib22,bib23,bib24} at first. Recent studies have indicated that common datasets contain a large number of easy examples and a small number of hard examples \cite{bib25}. The rareness of hard examples obstructs the enhancement of learning ability to network. There are three popular ways to deal with the problem: (1) Increase the number of hard examples within the acceptable cost of data acquisition. (2) Introduce data augmentation to enlarge the number of hard examples. (3) Design a loss function base on hard mining. As usual, the third method would be adopted because of the difficulty of acquisition of hard example. Classical hard mining methods are OHEM (Online Hard Example Mining) \cite{bib26}, MSML (Margin Example Mining Loss) \cite{bib27} , Focal Loss \cite{bib28}, etc. OHEM collected the regions of interest, ROIs, with relatively large loss from a set of candidate regions and then retrained it. \cite{bib29} draws on OHEM and the confidence of the Region Proposal Network is used to collect hard negative examples and send them to the network for retraining to improve the accuracy of target detection. In each batch, MSML will select the hardest positive example pair and the hardest negative example pair for backpropagation, so as to guide the network to learn the hard example. In the study of pedestrian re-recognition, the author also tries to solve the problem of hard examples by referring to MSML and proposes batch hard loss. In each batch, the hardest similar examples and dissimilar examples are selected as the punishment of the network to achieve better efficiency of detection \cite{bib30}. Focal loss is similar to OHEM, but it incompletely discards the punishment of easy examples. It only reduces the contribution of hard examples to the loss function and improves the penalty weight of hard examples.

There are some studies on deep supervised hash in hard examples problem. \cite{bib9} studied the problem by imposing a large penalty on similar images whose hamming distance is greater than the given radius threshold. They shorten the distance between similar examples in hamming space. Base on previous methods, Dual Hinge Loss is designed to imposes an exponential penalty on hard similar examples and boundary penalty on easy similar examples. The loss prevents easy examples from dominating in the optimization process \cite{bib10}. Except pick hard examples according to the hamming distance, PCDH \cite{bib31} extracts the positive example pair with the largest distance and the negative example pair with the distance less than the threshold in the feature space for backpropagation. The authors all above take into account the shortage of hard examples and select the hard examples through certain rules. Then they adjust the penalty weights of hard example pairs to achieve the purpose of guiding the network to learn the hard examples.

Although their method improves the accuracy of image retrieval, it does not give full play to the powerful learning ability of the network. First, the selection of hard examples uses the distance between discrete hash codes, whose computational process loses a lot of effective information. Additionally, they improve the ability of the network to learn the hard example only by modifying the penalty weights of the hard example. They ignore the influence of key feature, which do not fully reveal the efficacy of the learning hard example. Redundant information hinders the learning of hard examples. It leads to useless and even harmful feedbacks during backpropagation. Therefore, it is crucial to extract useful features from the network for learning hard examples.

In view of the above deficiencies, we propose HEGH to enhance the ability of learning hard examples from the two aspects of weak key features extraction and the shortage of hard examples.

\section{Hard Example Guided Hashing for Image Retrieval}\label{sec3}

Our goal is to improve the ability of the network to learn hard examples from two aspects: improve the ability of key feature extraction and alleviate the shortage problem of hard examples. We design a new network structure, guide the network to focus on the target region and learning the key features from hard examples. It helps generated hash code that contains more accurate semantic information. In addition, we design a hard pair-wise loss function to alleviate the imbalance proportion of examples. Finally, we reach an overall retrieval performance enhancement through hard examples learning.

\subsection{Architecture}

\begin{figure}[]
    \centering
    \includegraphics[width = \textwidth]{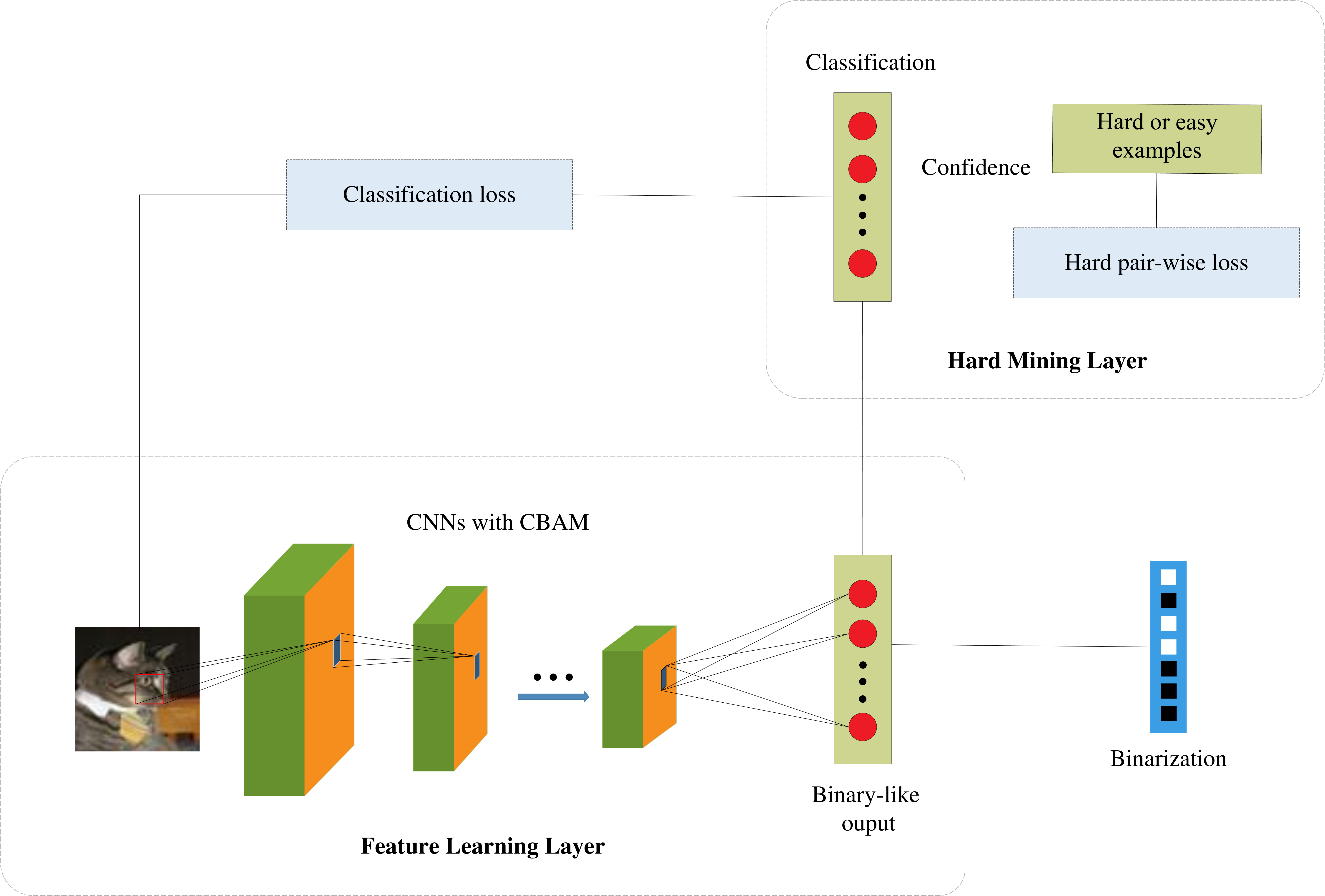}
    \caption{Model architecture of HEGH}\label{fig.2}
\end{figure}

Our network architecture of HEGH is shown in Fig. \ref{fig.2}, including feature learning layer and hard mining layer. Feature learning layers is a set of layers to extract features of the original image through CNNs with CBAM. The output of feature learning layer is binary-like output, mapped to the hash code. Hard mining layer is responsible for estimating the hard degree of examples relying on classification prediction. Same as some deep hashing methods \cite{bib17,bib32}, a full connection layer is responsible to predict label by learning semantic features. It is added at the end to assist binary-like output generates hash code including rich semantic information. In addition, the confidence of the classification layer, which reflects the classification ability of the network, is used to evaluate the hard degree of the example.

In this framework, it easily converts image to binary hash code. Firstly, the original image data is forward-propagated in feature learning part. Next, the network quantifies the binary-like output to get the hash code.

There are two parts in our loss function: classification loss and hard pair-wise loss. Classification loss measures classification errors. Hard pair wise loss picks hard example according to confidence and allocates different punishment weights for examples.

\subsection{Feature Learning Layer}

\subsubsection{CNNs with CBAM}

The existence of redundant information affects network extracting and learning high-level features in hard examples. By guiding the network to focus on the target region, CNNs with CBAM generates hash code containing more accurate semantic information and improves the learning of the hard examples.

At present, most of the deep hashing network structures adopt the common classification network, such as AlexNet \cite{bib33}, VGGNet \cite{bib34}, which effectively utilize the representation capability in the field of image classification. They have competitive performance compared with DSH \cite{bib35}, semi-manual CNNH \cite{bib18} and other early deep hashing. The features extracted by these classification networks are not all key features and may be mixed with redundant features because of the undifferentiated treatment of regions and channels.

The redundant feature influence is tiny for classification task, because the classification task does not require high feature quality. For image retrieval studies, the redundant features strongly impact the calculation of the similarity between images while order of similarity is counted into consideration. Furthermore, it increases the difficulty of learning key feature in hard examples due to the large portion of unnecessary information in the back-propagation.

\begin{figure}[!ht]
    \centering
    \includegraphics[width = 0.9\textwidth]{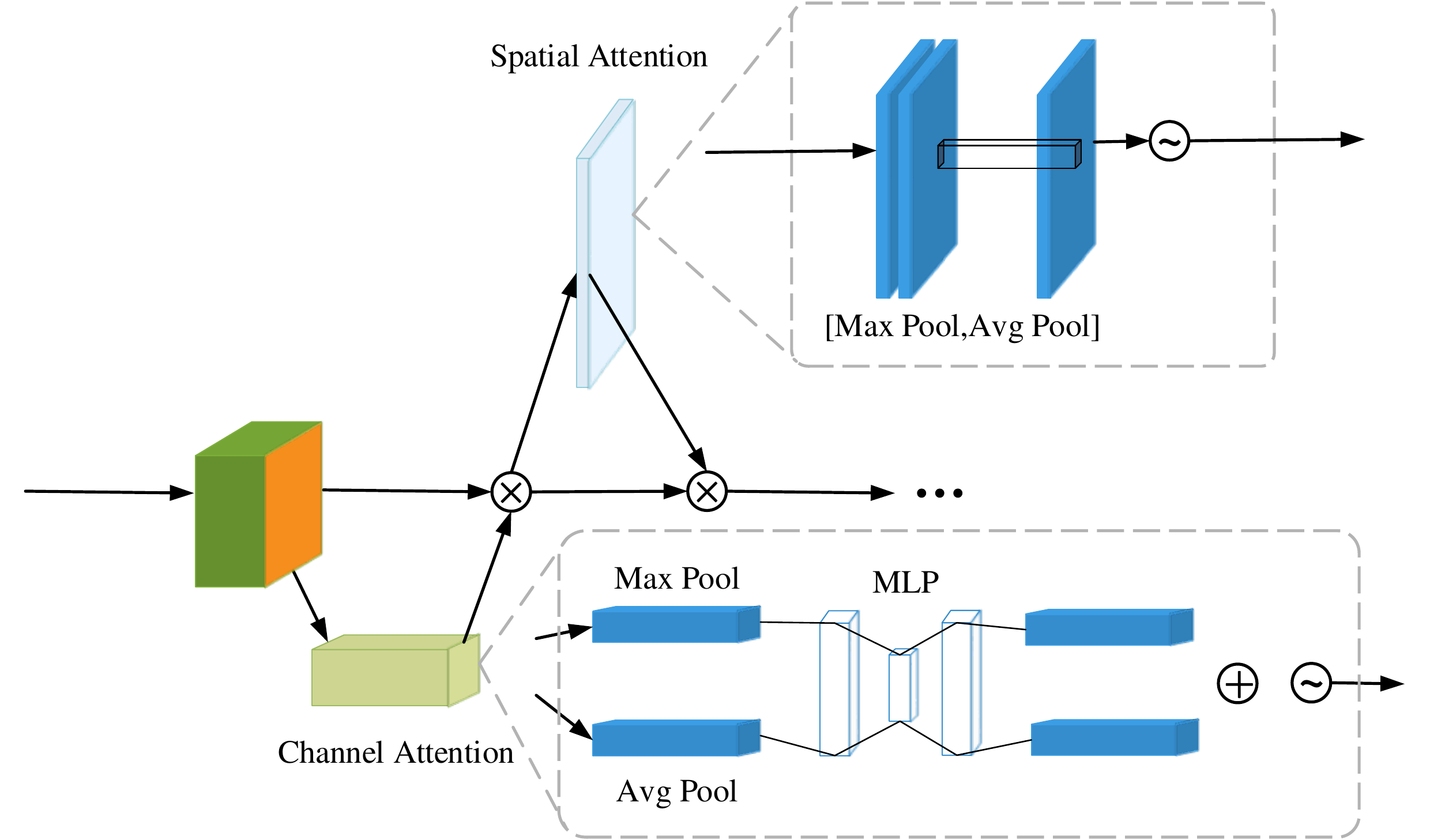}
    \caption{Detailed diagram of the CBAM}\label{fig.3}
\end{figure}

In order to help the network extract and learn key features, we design a network, CNNs with CBAM. CBAM (Convolutional Block Attention Module) \cite{bib36} combines channel attention and spatial attention to enable the network to learn which features and areas are important, so as to learn better target features. The structure of CBAM is shown as Fig. \ref{fig.3}. The channel attention adopts the max pool and average pool in the shared network to get the focus channel. The spatial attention converges the max pool and average pool along the channel axis, and forwards them into the convolutional layer to obtain the focus spatial region. With the help of channel attention and spatial attention, the network identifies important channels and regions by self-learning, which reduces the extraction of redundant features. This method avoids the key features being buried and enhances the ability of learning hard examples in image retrieval.

\begin{figure}[]
    \centering
    \includegraphics[width = \textwidth]{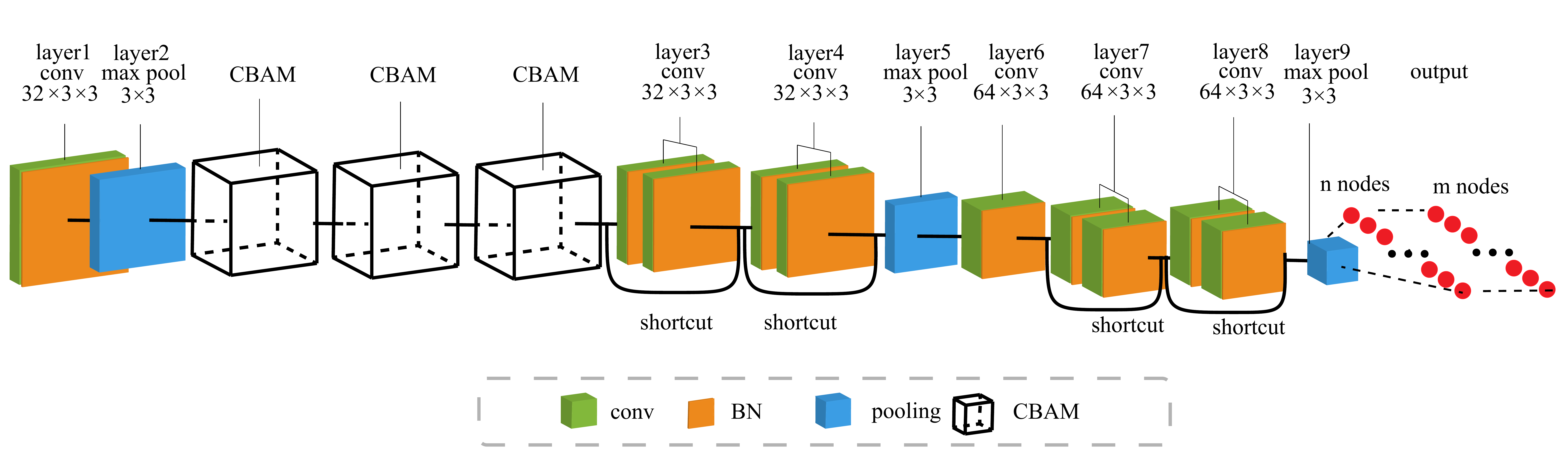}
    \caption{Network architecture of the CNNs with CBAM proposed in this paper.}\label{fig.4}
\end{figure}

Through a large number of design experiments, we get the final network, CNNS with CBAM, as shown Fig. \ref{fig.4}. During the training of CNNS with CBAM, CBAM guides the network to focus on the target of the image and effectively extract key feature.

\subsubsection{Design details}

\begin{figure}[]
    \centering
    \includegraphics[width = \textwidth]{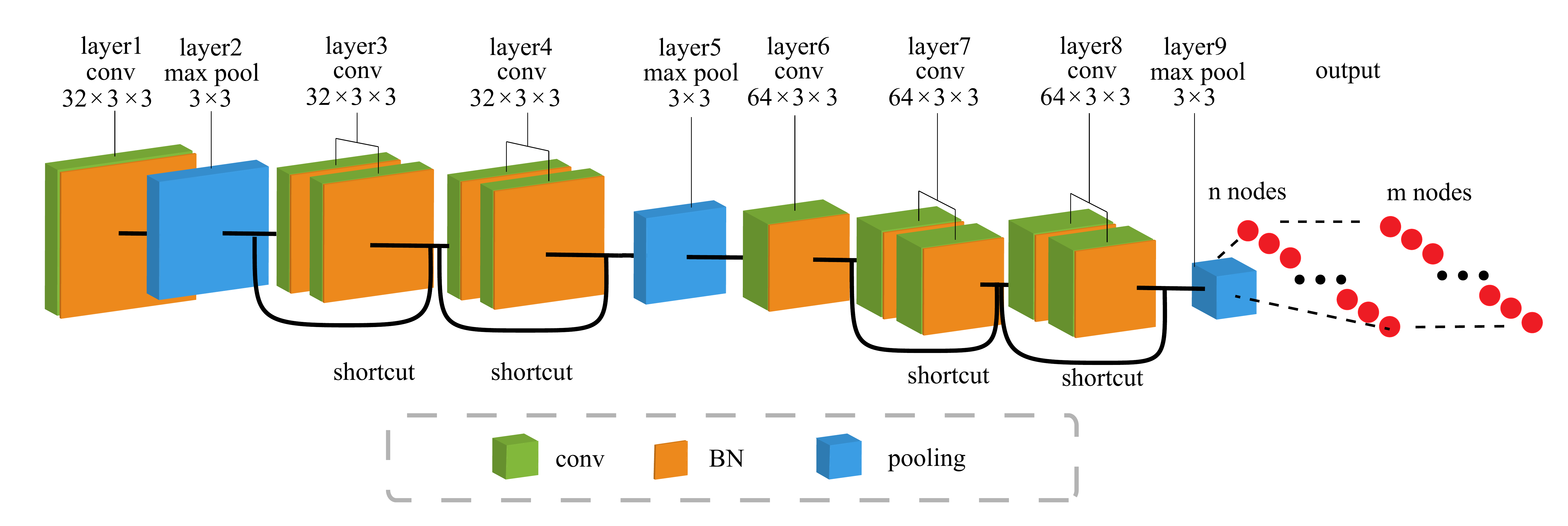}
    \caption{Basic CNN network architecture}\label{fig.5}
\end{figure}

This part mainly discusses the network design experiment. In the process of network design, based on Fig. \ref{fig.5}, we explored the location and number of CBAM modules placed in the network through experiments. Secondly, we added shortcut connections and prove their importance by experiment to avoid the problem of gradient disappearance caused by increasing network depth. The rest are devoted to discuss the special design of the network. The experimental results quoted here are all based on CIFAR-10 dataset, and the model accuracy of 12 bits is reported using the detection accuracy of the test dataset.

\begin{enumerate}
    
    \item [(1)] CBAM
     
Firstly, for location, CBAM module will be placed behind layer2, layer4, layer8 of the basic network shown in Fig. \ref{fig.5}, for experiments. It is found in Table. \ref{table.1} that the further CBAM is placed, the lower MAP of the model, 0.022 and 0.0417 lower than that placed in the front respectively. This phenomenon is related to the size of the input. Due to the resolution of CIFAR-10 is only 32*32, the image becomes very small after each down sampling. When the image is very small, the effect of CBAM is very weak.

Then, we combined the CBAM module in different positions, and MAP is worse (0.0152 $\sim$ 0.0385) than one CBAM module placed in the front. So, adding one CBAM module after layer2 alone is better than hybrid combinations. Therefore, we put the CBAM module behind layer2 of the network.
    
\begin{table}[]
    \begin{center}
    \begin{minipage}{\textwidth}
    \caption{Result of placing CBAM in different location}\label{table.1}
    \begin{tabular*}{\textwidth}{@{\extracolsep{\fill}}lllllll@{\extracolsep{\fill}}}
    \toprule
    & \multicolumn{6}{c}{Location of CBAM}\\
    \midrule
    &Layer2 & Layer4 & Layer8 & \makecell[l]{Layer2+\\Layer4}&\makecell{Layer4\\+Layer8}&\makecell{Layer2\\+Layer4\\+Layer8}\\
    \midrule
    MAP&\textbf{0.869}&0.847&0.827&0.865&0.853&0.830\\
    \botrule
    \end{tabular*}
    \end{minipage}
    \end{center}
\end{table}

Next, we experiment the impact of the number of CBAM module stacks behind layer2 on the retrieval capability of the model. We stacked two or three CBAM modules respectively. According to Table. \ref{table.2}, three CBAM modules are finally stacked behind layer2 of the network. According to the experiment, the more CBAM does not mean the better improvement. Because the output of CBAM is [0,1], the eigenvalue will be smaller and smaller after multiple times of product.

\begin{table}

    \begin{minipage}[t]{.2\textwidth}
        \setlength{\abovecaptionskip}{0pt}%
        \caption{Result of different number of CBAM are placed behind Layer2}\label{table.2}
    \end{minipage}\hfill
    \begin{minipage}[t]{.75\textwidth}\vspace*{0pt}%
        \begin{tabular*}{\textwidth}{@{\extracolsep{\fill}}lllll@{\extracolsep{\fill}}}
            \toprule
            & \multicolumn{4}{c}{Number of stacks}\\
    \midrule
    &One&Two&Three&Four\\
    \midrule
    MAP&0.869&0.856&\textbf{0.879}&0.856\\
            \bottomrule
        \end{tabular*}
    \end{minipage}

\end{table}

    \item [(2)] Residual shortcuts
    
To assess the importance of shortcut connections in the network, we removed the connections from layers of layer3, layer4, layer7, layer8 and observed the change in MAP. In Table. \ref{table.3}, we find that the decrease of MAP is about between 0.022 and 0.049 when we respectively removed shortcut connections to each of the four locations. Before the removement, the performance gain is because short connections resolve the disappearance of gradients caused by the number of layers increases.

\begin{table}

    \begin{minipage}[t]{.2\textwidth}
        \setlength{\abovecaptionskip}{0pt}%
        \caption{Result of removing different shortcuts}\label{table.3}
    \end{minipage}\hfill
    \begin{minipage}[t]{.75\textwidth}\vspace*{0pt}%
        \begin{tabular*}{\textwidth}{@{\extracolsep{\fill}}lllll@{\extracolsep{\fill}}}
            \toprule
            &\multicolumn{4}{c}{Different shortcut removeds} \\
            \midrule
            &Layer3&Layer4&Layer7&Layer8\\
            \midrule
            MAP&0.851&0.857&0.830&0.850\\
            \bottomrule
        \end{tabular*}
    \end{minipage}

\end{table}

According to previous experiments, we obtained the final network structure, as shown in Fig. \ref{fig.4}
    
\end{enumerate}

\subsection{Hard Mining Layer}

Generally, the number of hard examples in common dataset is much lower than the number of easy examples, and this problem is more serious in single batch at training. The shortage of hard examples weakens the ability of learning hard examples. For the network, easy examples mean easy to learn. If we don't increase the proportion of hard examples in back propagation, valid information will be sparse in the later epoch of training. Consequently, we try to guide the network to learn hard examples by treating the disadvantage in the number.

For general loss functions, easy examples constituted the significant part of the loss and dominated the gradient of back propagation \cite{bib28}. Original pair-wise loss function is shown in Eq.\ref{Eq.1}. $b_1$ and $b_2$ are the corresponding binary-like outputs of a pair of images respectively. We define $Y$=0 if the pair of images is similar, and $Y$=1 otherwise. The first term of Eq.\ref{Eq.1} penalizes similar images of different hash codes, and the second penalizes dissimilar images of close hash codes when their hamming distance is below the marginal threshold $m_c$. $m_c$ is a manually hyperparameter.

\begin{align}
    L_{pair}(b_1, b_2, Y)=&\frac{1}{2}(1-Y)\lVert b_1-b_2\rVert^{2}_{2}+\frac{1}{2}Ymax(m_c-\lvert\lvert b_1-b_2\lvert\rvert)^{2}_{2}, 0) \label{Eq.1}\\
    &s.t.b_j\in {+1, -1}, j\in{1, 2} \nonumber
\end{align}

Original pair-wise loss function does not update the punishment weights of each example. Due to the superiority of easy examples in quantity, easy examples dominate the back propagation of each batch. The network quickly learns the information of an easy example pair and the large part of the information is worthless in the later training. This easily leads to the dissimilar example with certain similar coarse-grained outlooks will be considered similar. 

We try to avoid the problem by alleviating the imbalances in quantity. Because hard examples are difficult to obtain, we adjust the penalty weights of examples to solve the problem or collect hard examples for retraining. To keep training speed, we design a new pair-wise loss function to update the penalty weights. There are two steps: the selection of hard examples and penalty weights modification.

\begin{enumerate}

\item [(1)] \textbf{The selection of hard examples}
    
The first step of hard mining is to select hard examples. At present, image retrieval is mainly based on hamming distance or feature distance to select hard examples. The hamming distance is calculated by discrete hash code. The computational process leads to the loss of features, so that the selection of hard examples is inaccurate. Feature distance is a relative information, which cannot well represent the semantic understanding ability of the network. It cannot obviously show that which image the model mistakes in the pair. Therefore, our selection of hard examples relies on the distance between category prediction and the ground-truth label. A better selection of hard examples will come out. 

 We introduce a classification layer to predict label and a classification loss into the loss function to conduct the classification task training. It is necessary to ensure that the network has certain classification ability since the selection of hard examples needs to be based on the confidence of the classification layer. The formula for classification loss is shown below:

\begin{equation}
    L_{class}=\sum_{i=1}^{N}\sum_{j=1}^{N_c}  \lVert y_{i,j}-p_{i,j}\rVert^2_2 \label{Eq.2}
\end{equation}

\noindent where $y_{i,j}$ is the ground-truth label vector, $p_{i,j}$ is the model’s estimated probability for every label, $N_c$ represents the number of categories and $N$ is the number of the training set.

According to the concept of hard examples, in the case of correct prediction, the confidence is negatively correlated with the hard degree of examples. On the contrary, the confidence is positively correlated with the hard degree of examples. We assume that $P$ is the hard degree of the example, and $p$ represents the confidence from output of the classification layer. The relationship between $P$ and $p$ is shown as Eq.\ref{Eq.3}. When it is a similar example pair, $p$ is inversely proportional to $P$. The small value of $p$ means that the pair is considered as a hard similar pair. When it is a dissimilar example, $p$ is in proportion to $P$. The high value of $p$ means that the pair is considered as a hard dissimilar pair.

\begin{equation}
    P=\begin{cases}
        p, &if\quad Y=1\\
        1-p, &otherwise\\
    \end{cases} \label{Eq.3}
\end{equation}

\item [(2)] \textbf{Penalty weights modification}
    
The second step of hard mining is updating penalty weights. Drawing on the idea of Focal Loss\cite{bib28}, a novel pair-wise loss function is designed. In each batch, we automatically reduce the penalty weights of easy examples and increase the weights of hard examples. The punishment weights of example are determined by $P$. Our pair-wise loss function is shown as Eq.\ref{Eq.4}. There are $N_p$ training pairs randomly selected from the training images. The hard degree of examples is proportional to the penalty weights. The hyperparameter $\gamma$ is the rate of downward weighting of the easy example. When $\gamma$=0, the loss function is original pair-wise loss function. The larger $\gamma$ is, the smaller the penalty for the easy example is, with an exponential downward weighting.

\begin{equation}
    L_{hard}=\sum^{N_p}_{i=1}(P_{i,1}P_{i,2})^{\gamma}L_{pair}(b_{i,1},b_{i,2},Y_i) \label{Eq.4}
\end{equation}

Image retrieval usually adds a regularizer to the output of the network, and imposes a limit on the binary-like layer to make its output approximate (-1, +1). We use L1-norm as the regularizer, because the computation of L1-norm is relatively small, which accelerates the training process of the network. At the end, our hard pair-wise loss is shown as:

\begin{align}
    L^{'}_{hard}=&\sum^{N_p}_{i=1}\{ (P_{i,1}P_{i,2})^{\gamma}L_{pair}(b_{i,1},b_{i,2},Y_i) \nonumber\\
    &+\lambda(\lVert\ \lvert b_1 \rvert - sgn(b_1)\ \rVert_1+\lVert\  \lvert b_2 \rvert - sgn(b_2) \ \rVert_1)\} \label{Eq.5}
\end{align}

\noindent where $lambda$ is the regularization parameter.

We combine the losses as follows:

\begin{equation}
    L=L^{'}_{hard}+\mu L_{class} \label{Eq.6}
\end{equation}

$L^{'}_{hard}$ is responsible for optimizing the hash code and solving the shortage of hard examples. $L_{class}$ improves the semantic understanding ability of the network to optimize the selection of hard examples.

\end{enumerate}

\section{Experiment}\label{sec4}

\subsection{Datasets and metrics}

We evaluate the effectiveness of Hard Example Guided Hashing for Image Retrieval, HEGH, by against several hashing-based image retrieval methods on two standard benchmark datasets: CIFAR-10 and NUS-WIDE.

CIFAR-10 is a public image dataset, containing 60,000 small images of 32×32, divided into 10 mutually exclusive categories. We randomly select 50,000 images for the training set and 10,000 for the testing set, with the same number for each category.

NUS–WIDE is a multi-tag image dataset created by the Media Search Lab at the National University of Singapore. The dataset consists of 269,648 images, divided into 81 human-labeled real categories, and associated with 1,000 of the most frequent labels. Following the experimental protocols in DSH\cite{bib35}, 21 most frequent labels are collected in our experiments. Each label associates with at least 5000 images. Therefore, we used 195,834 images warped to 64×64, 10000 as the testing set and the rest is used as the training set.

Three widely metrics \cite{bib6} are mainly used in the experiment: Mean Average Precision (MAP) and Precision of the top k (Precision@k), Precision Within Hamming Radius 2 (P@H$\leq$2).

\begin{itemize}
    \item Mean Average Precision (MAP): It is calculated as the mean AP of each query image in the Hamming-distance ranking list.
    
    \begin{equation}
        MAP=\frac{1}{Q}\sum^{Q}_{q}AP(q)\label{Eq.7}
    \end{equation}

    \begin{equation}
        AP(q)=\frac{1}{N}\sum^{N}_{i=1}\frac{i}{position(i)}\label{Eq.8}
    \end{equation}

\noindent where $Q$ is the numbers of query sets, $N$ represents the number of examples similar to the query image $q$, $position(i)$ represents the position of the similar example $i$ in the retrieval result list. Higher MAP means a better retrieval result.

    \item Precision@k of the Ranked List (Precision@k): It is computed as the precision for the query image in top k closest result sorted by Hamming-distance.
    
\begin{flalign}
    Precision@k=\frac{1}{k}\sum^{k}_{i=1}rel(i) \label{Eq.9}
\end{flalign}

\noindent where $rel(i)$ is the similarity between the query and the ith ranked image of the ground truth and $rel(i)=1$ if the query is similar.

    \item Precision Within Hamming Radius 2 (P@H$\leq$2): The retrieval results of the query images are sorted by the Hamming distance. We calculate the mean precision when the hamming distance between the query image and the database image is less than or equal to 2.
    
    \begin{equation}
        P@H\leq2=\frac{1}{N_H}\sum^{N_H}_{i=1}rel(i) \label{Eq.10}
    \end{equation}
    
\noindent where $N_H$ is the number of examples whose hamming distances are less than or equal to 2.

\end{itemize}

These three evaluation protocols measure the performance of the hash methods\cite{bib37}. The greater the value of these evaluation protocols, the better the performance.

\subsection{Experiment setup}

Like most deep hashing algorithms, in our experiments, similarity labels are defined by semantic-level category labels. The similarity criterion for the dataset of a single label is that two image labels are the same and the images can be considered to be similar. According to the previous multi-label image similarity standard, two images are considered similar if they share at least one semantic label.

In order to better verify the performance of our method, we set up six experiments:

\begin{itemize}[(1)]
    \item We set several parameters with different values to test the influence of parameter values on the image retrieval model. The main parameter affecting hard mining is discussed in this part. It proves the importance of solving the shortage of hard examples;
    \item In order to better explore the contribution of CBAM to the method, we compare the retrieval accuracy of our network and that without CBAM on the CIFAR-10, and visualized the areas of key features. It verifies the capability and the optimization effect of key feature extraction;
    \item We set up comparison experiments on CIFAR-10 and NUS-WIDE to evaluate the overall performance of our method compared with other retrieval models under different hash code bits.
    \item The accuracy of the top k retrieval results is consistent with the performance of the image retrieval model in practical application. Therefore, we further explore the top k retrieval results of our model;
    \item In the field of image retrieval, researchers are pursuing the generation of more compact hamming distance to reduce the length of hash code. We test the accuracy within the hamming radius 2 to see how compact we generate the hash code;
    \item Finally, we visualize the distribution of hash codes to observe the overall situation of the generated hash codes.
\end{itemize}

\subsection{Results}

\subsubsection{Parameter analysis}

In this part, we validate the effectiveness of the hard pair-wise loss function. There are four hyperparameters in the proposed method, $\lambda$, $m_c$, $\mu$ and $\gamma$. $\lambda$ is used to adjust the regularization term and $m_c$ is the marginal threshold in loss function. The value of $\lambda$ and $m_c$ is taken from DSH \cite{bib35}. $m_c$ is set $2k$ ($k$ is hash code number) to encourage the code of dissimilar examples to differ at no less than $\frac{k}{2}$ bits. $\mu$ represents the proportion of Classification loss. Referring to \cite{bib17}, $\mu$ is set to 1. $\gamma$ is used to control the influence degree of weight modification. The value settings of $\lambda$, $m_c$ and $\mu$ are based on the former study works by other researchers. Our experiment mainly discusses the parameter $\gamma$ that affects hard mining.

Table \ref{table.4} shows the results obtained by using different values of $\gamma$ when $\lambda$ is 0.01 and $\mu$ is 1. When $\gamma$=0, it is equivalent to the original pair-wise loss function. We make three observations from the results. First, the best results are obtained when $\gamma$=1, whose MAP is 5\% higher than the value of original pair-wise loss. Therefore, hard mining effectively improves the retrieval precision. Second, when $\gamma > $1, $\gamma$ and Map are negatively correlated. If the value of $\gamma$ is too large, we find that the penalty weights of the pairs does not differ significantly, so hard mining work improperly. Third, it is not a good idea to adjust the penalty weight too excessive, which seriously narrows the difference of hamming distance between the pairs. The penalty weight is supposed to be modified within an appropriate range, to preserve the difference of hamming distances.

The validity of hard mining proves that the retrieval performance can be improved effectively by solving the shortage of hard examples.

\begin{table}[]
    \begin{center}
    \begin{minipage}{\textwidth}
    \caption{Results of MAP for different parameter value of $\gamma$ on CIFAR-10 when the bit is 12.}\label{table.4}
    \begin{tabular*}{\textwidth}{@{\extracolsep{\fill}}cccccccc@{\extracolsep{\fill}}}
    \toprule
    &$\gamma$=0 & $\gamma$=0.5 & $\gamma$=1 & $\gamma$=1.5 & $\gamma$=2 & $\gamma$=2.5 & $\gamma$=3 \\
    \midrule
    MAP&0.822&0.823&\textbf{0.871}&0.820&0.826&0.815&0.762\\
    \botrule
    \end{tabular*}
    \end{minipage}
    \end{center}
\end{table}

\subsubsection{Effectiveness of CBAM}

We remove CBAM from the proposed network and test MAP without CBAM in CIFAR-10. The addition of CBAM increases the MAP (0.054-0.070), shown in Table \ref{table.5}. Thus, CBAM effectively improve the retrieval performance.

\begin{table}

    \begin{minipage}[t]{.3\textwidth}
        \setlength{\abovecaptionskip}{0pt}%
        \caption{Results of MAP for proposed network without CBAM and proposed network in CIFAR-10.}\label{table.5}
    \end{minipage}\hfill
    \begin{minipage}[t]{.65\textwidth}\vspace*{0pt}%
        \begin{tabular*}{\textwidth}{@{\extracolsep{\fill}}ccccc@{\extracolsep{\fill}}}
            \toprule
            &12-bit&24-bit&36-bit&48-bit \\
            \midrule
            HEGH without CBAM&0.812&0.820&0.807&0.814\\
            \midrule
            HEGH&\textbf{0.871}&\textbf{0.874}&\textbf{0.877}&\textbf{0.869}\\
            \botrule
        \end{tabular*}
    \end{minipage}

\end{table}

\begin{figure}[]
    \centering
    \includegraphics[width = \textwidth]{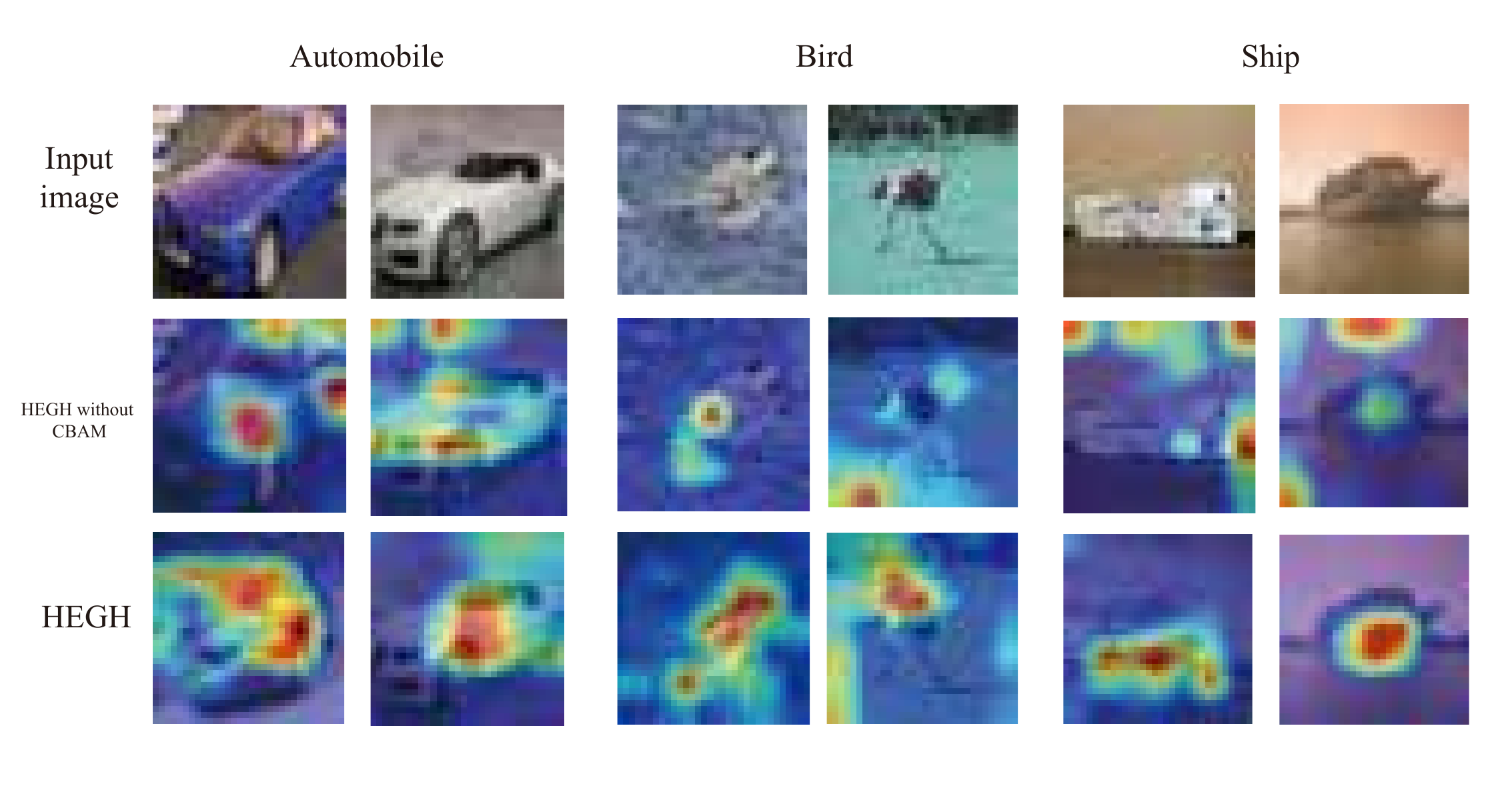}
    \caption{Grad-CAM\cite{bib37} visualization results. We compare the visualization results of basic CNN without CBAM and our final network. The true label is shown above each input image.}\label{fig.6}
\end{figure}

In order to understand whether CBAM helps the network to improve the retrieval performance, we adopt Grad-CAM \cite{bib32} visualization method to clearly show the area of network concern. Fig. \ref{fig.6} indicates that CBAM assists network to focus on the region that are important for predicting a class. The network makes good use of the target information and integrate the key features. Focusing on key areas is conducive to reducing the extraction of redundant information and spending more resource on learning high-level features. It improves the ability of key features extraction and the generation hash code of hard examples with precise semantic information.

\subsubsection{MAP for different methods}

Table \ref{table.6} and Table \ref{table.7} show MAP on the CIFAR-10 and NUS-WIDE datasets at different bits. There are four different lengths of hash code in this experiment (i.e., from 12-bit to 48-bit). Due to the variety of experimental settings of existing methods, the comparison methods of each dataset are inexactly same, and some data are missing. From the results, the method proposed in this paper has better results than other methods. HEGH is superior in both single-label and multi-label datasets. For CIFAR-10, with the increase of bit, the value of map decreases. It may be because existing bits in the generated hash codes are redundant or even harmful to the retrieval accuracy. In future work, we will consider more about some improved strategies which are able to improve generated hash code.

\begin{table}

    \begin{minipage}[t]{.2\textwidth}
        \setlength{\abovecaptionskip}{0pt}%
        \caption{Results of MAP for proposed network without CBAM and proposed network in CIFAR-10.}\label{table.6}
    \end{minipage}\hfill
    \begin{minipage}[t]{.75\textwidth}\vspace*{0pt}%
        \begin{tabular*}{\textwidth}{@{\extracolsep{\fill}}ccccc@{\extracolsep{\fill}}}
            \toprule
            Methods&12-bit&24-bit&36-bit&48-bit \\
            \midrule
            \textbf{HEGH}&\textbf{0.871}&\textbf{0.874}&\textbf{0.877}&\textbf{0.869}\\
            \midrule
            DSH(2016)\cite{bib16}&0.616&0.651&0.661&0.676\\
            \midrule
            DSDH(2017)\cite{bib17}&0.740&0.786&0.801&0.820\\
            \midrule
            SSAH(2019)\cite{bib38}&0.702&0.729&-&0.7513\\
            \midrule
            DBDH(2020)\cite{bib39}&-&0.736&0.742&0.748\\
            \midrule
            LiNet(2020)\cite{bib40}&0.740&0.786&0.801&0.820\\
            \midrule
            RODH(2020)\cite{bib41}&-&0.851&-&0.862\\
            \midrule
            CSH(2021)\cite{bib42}&-&0.822&-&0.827\\
            \botrule
        \end{tabular*}
    \end{minipage}

\end{table}

\begin{table}

    \begin{minipage}[t]{.2\textwidth}
        \setlength{\abovecaptionskip}{0pt}%
        \caption{Results of MAP for different numbers of bits on NUS-WIDE.}\label{table.7}
    \end{minipage}\hfill
    \begin{minipage}[t]{.75\textwidth}\vspace*{0pt}%
        \begin{tabular*}{\textwidth}{@{\extracolsep{\fill}}ccccc@{\extracolsep{\fill}}}
            \toprule
            Methods&12-bit&24-bit&36-bit&48-bit\\
            \midrule
            \textbf{HEGH}&\textbf{0.826}&\textbf{0.820}&\textbf{0.824}&\textbf{0.841}\\
            \midrule
            DSH(2016)&0.560&0.578&0.581&0.588\\
            \midrule
            DSDH(2017)&0.685&0.710&0.719&0.727\\
            \midrule
            SSAH(2019)&-&-&0.640&0.645\\
            \midrule
            DFH(2019)\cite{bib43}&0.795&0.780&0.804&0.812\\
            \midrule
            LiNet(2020)&0.776&0.808&0.820&0.829\\
            \midrule
            DBDH(2020)&0.814&0.820&0.801&0.809\\
            \botrule
        \end{tabular*}
    \end{minipage}
    
\end{table}

\subsubsection{Top retrieval results}

\begin{figure}[]
    \centering
    \includegraphics[width = \textwidth]{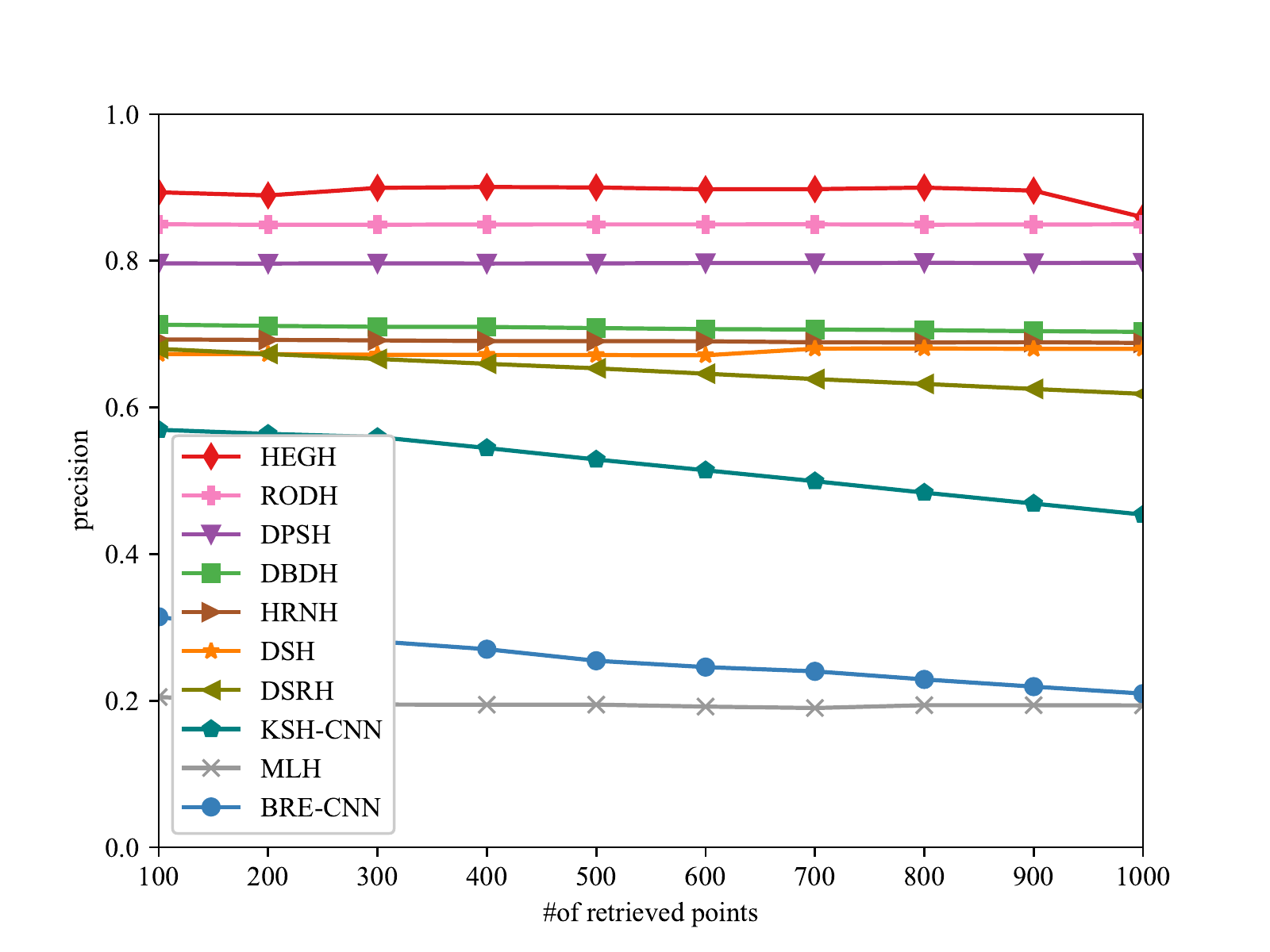}
    \caption{ Precision curves with 64 hash bits.}\label{fig.7}
\end{figure}

\begin{figure}[H]
\centering
    \subfigure[]{
    \label{fig.8.a}
    \includegraphics[width=0.3\textwidth]{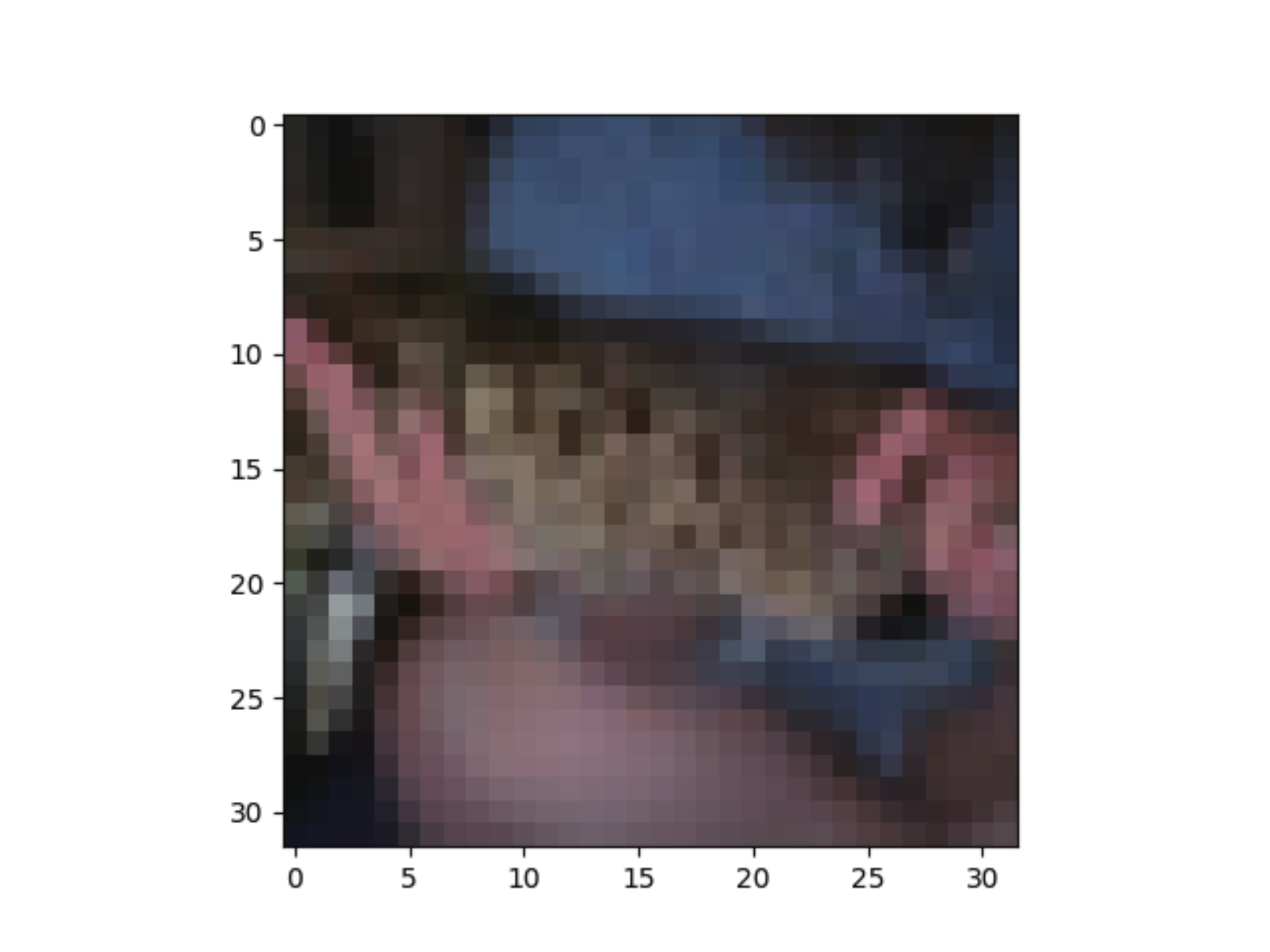}}
    \subfigure[]{
    \label{fig.8.b}
    \includegraphics[width=0.3\textwidth]{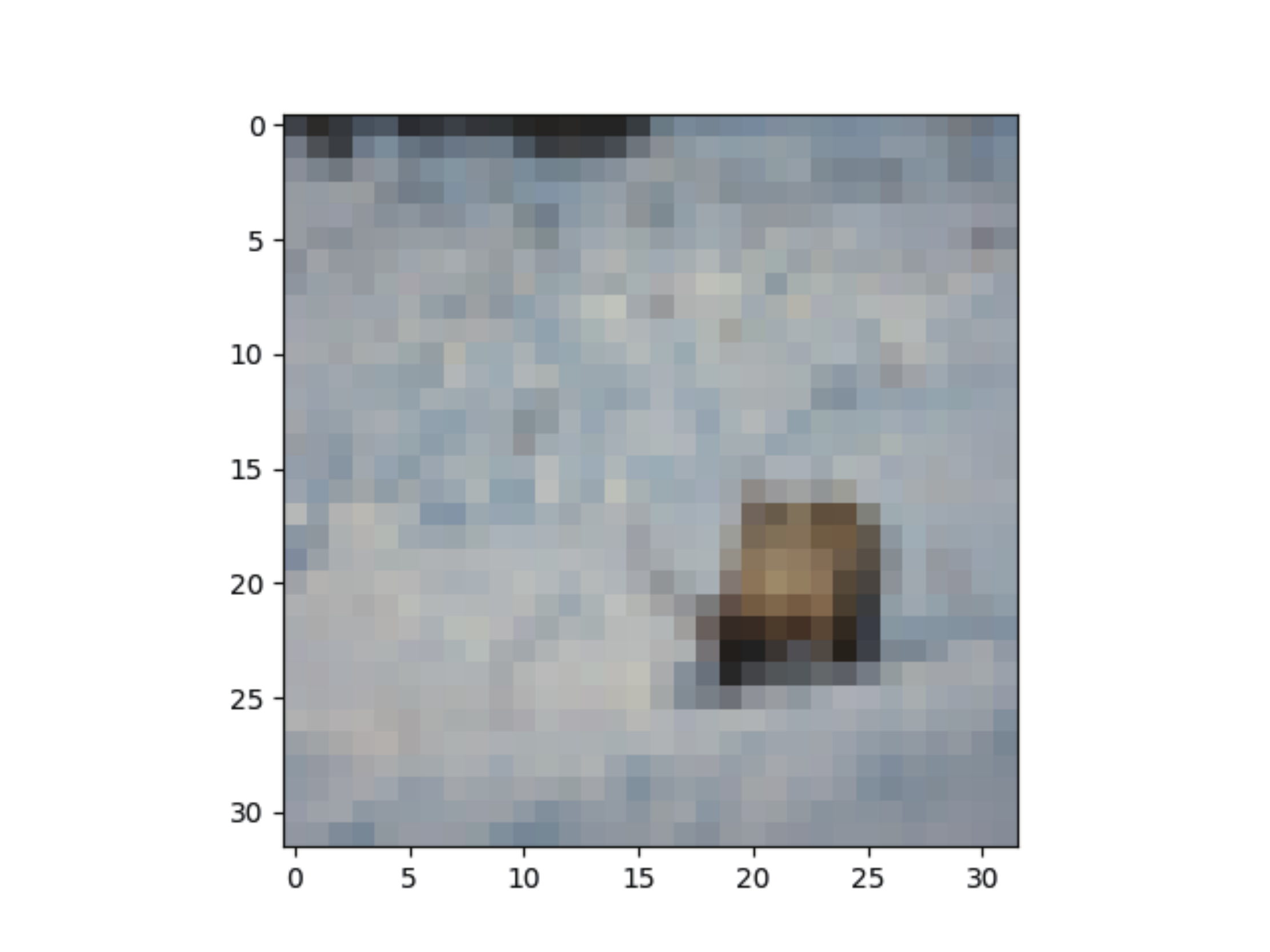}}
    \subfigure[]{
    \label{fig.8.c}
    \includegraphics[width=0.3\textwidth]{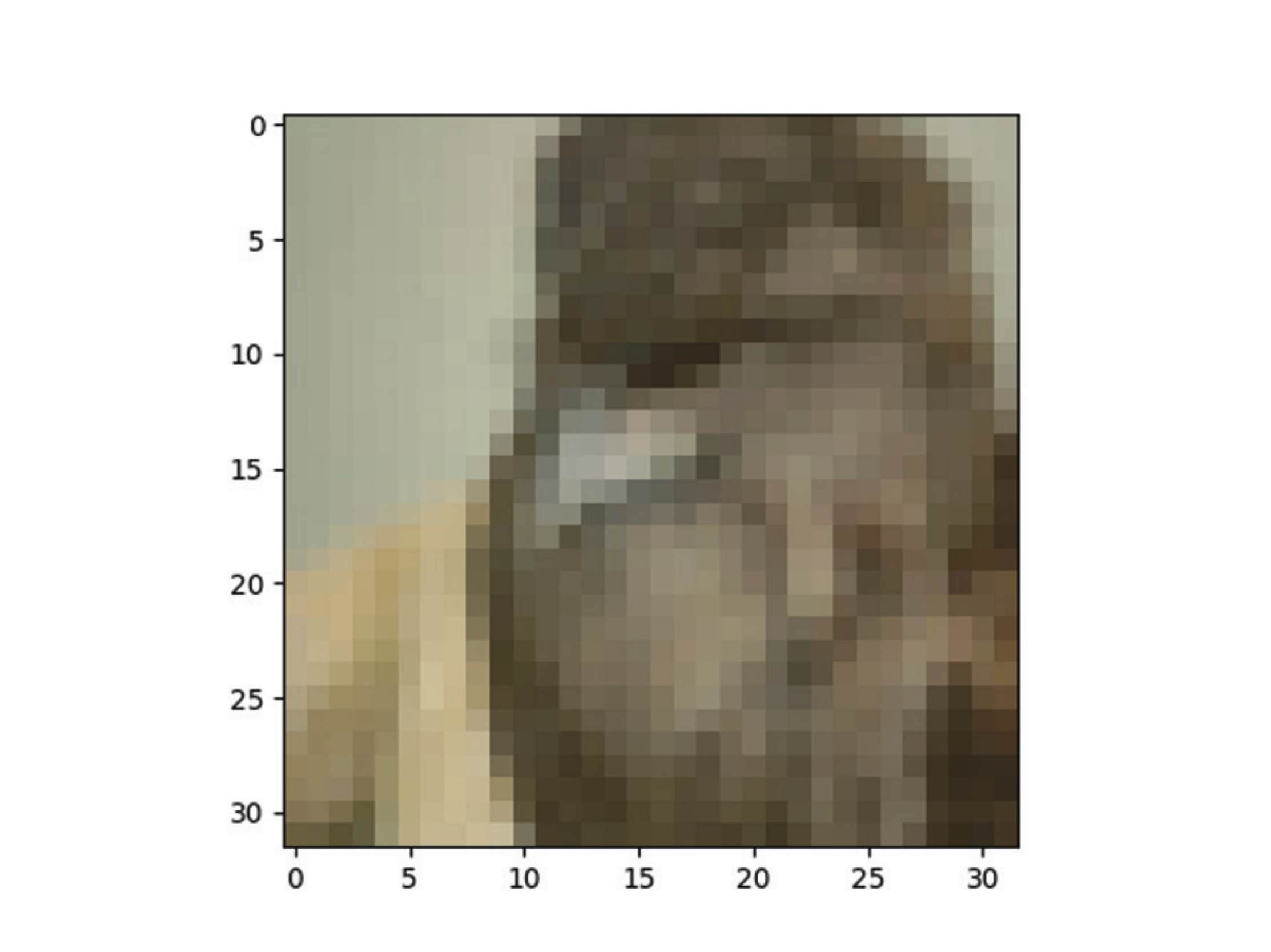}}
\caption{Error retrieval result}
\label{fig.8}
\end{figure}

Precision@k stands for relevancy of ranking results and is an important metric for evaluating retrieval systems. Precision@k is positively associated with the number of similar examples in the top k. Precision@k of different methods \cite{bib15,bib20,bib35,bib39,bib41,bib44,bib45,bib46,bib47} are listed in Fig. \ref{fig.7}. HEGH still shows better performance than existing methods. We find that the top 1000 result is only slightly better than existing methods. We looked at the images that are wrongly retrieved and find that the examples lacked many important features due to small target or low resolution. For example, in Fig. \ref{fig.8.a}, the cat's face is obscured, and many discriminative features are lost. The target area is small and the occlusion phenomenon is serious in Fig. \ref{fig.8.b}. For Fig. \ref{fig.8.c}, because of its low resolution, it is difficult for human eyes and network to recognize its semantics. These special examples make it hard for the network to extract features and our method is unable to handle them well.

\begin{figure}[]
    \centering
    \includegraphics[width = \textwidth]{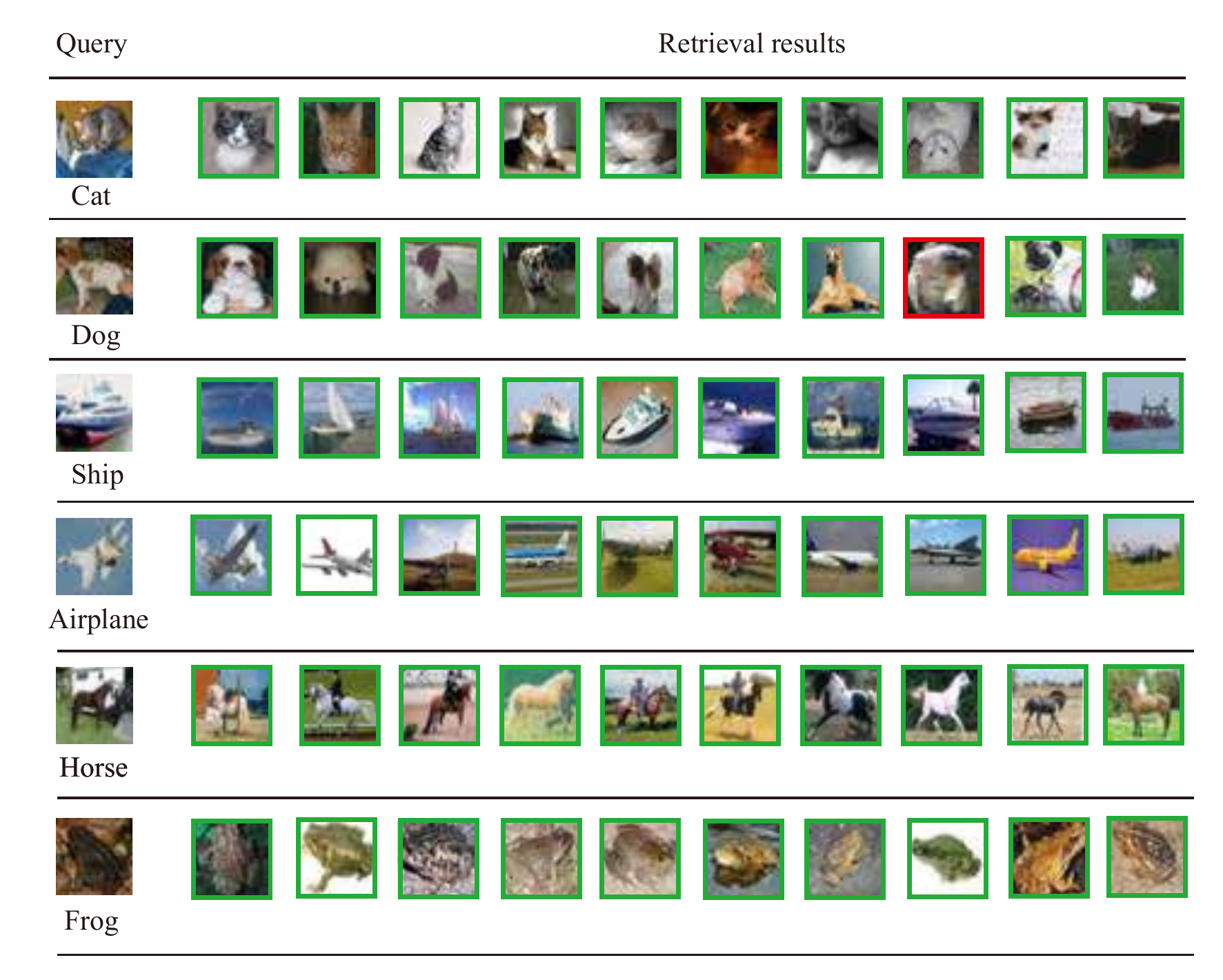}
    \caption{Top 10 retrieved images from CIFAR-10 dataset by our method. The first column is the query images. In the right columns, the red box represents the false retrieval image and the green box marks the true retrieval image.}\label{fig.9}
\end{figure}

Fig. \ref{fig.9} shows a partial retrieval of examples retrieved and sorted by ascending hamming distance. The green box indicates that similar images are successfully retrieved, while the red box indicates the retrieved image is dissimilar to the query image. According to the top retrieval results, we obtain a comparatively accurate ranking of retrieval results. It can be seen that CIFAR-10 has a large intra-class difference. HEGH still correctly retrieves similar images with consistent semantics. HEGH increases the network's attention to hard examples and helps the network find high-level features with more distinguishing. Therefore, HEGH precisely extracts the semantics of images and judge the similarity of images more exactly.

\subsubsection{Precision within hamming radius 2 (P@H$\leq$2)}

We evaluated the proposed HEGH and compared it with other methods at precision within Hamming radius 2. The hamming ranking only requires $O(1)$ time cost for each query, which enables really fast pruning.\cite{bib48} P@H$\leq$2 is important for image retrieval, because image retrieval want to use a shorter hash code to represent the example distance in the feature space. The higher P@H$\leq$2 proves that the generated hash code is more compact. \cite{bib35}The experimental results within the scope of CIFAR-10 are shown in Fig. \ref{fig.9}. We can see that our method achieves the best performance for each bit under precision within hamming radius 2. By increasing the penalty for hard example pairs, we produce the more compact hash code of hard similar example pairs, so P@H$\leq$2 is higher than other methods.

Each hamming distance is assigned less information in long codes than in shorter codes, so long hash codes are harder to compact. However, compared to other methods, as the bits of hash code increases, our precision does not decrease significantly, which reflects the optimization effectiveness of proposed method for hamming ranking. This may because the network extracts fewer redundant features and each bits contains more accurate information. Therefore, in the case of a long hash code, our performance is better.

\begin{figure}[]
    \centering
    \includegraphics[width = \textwidth]{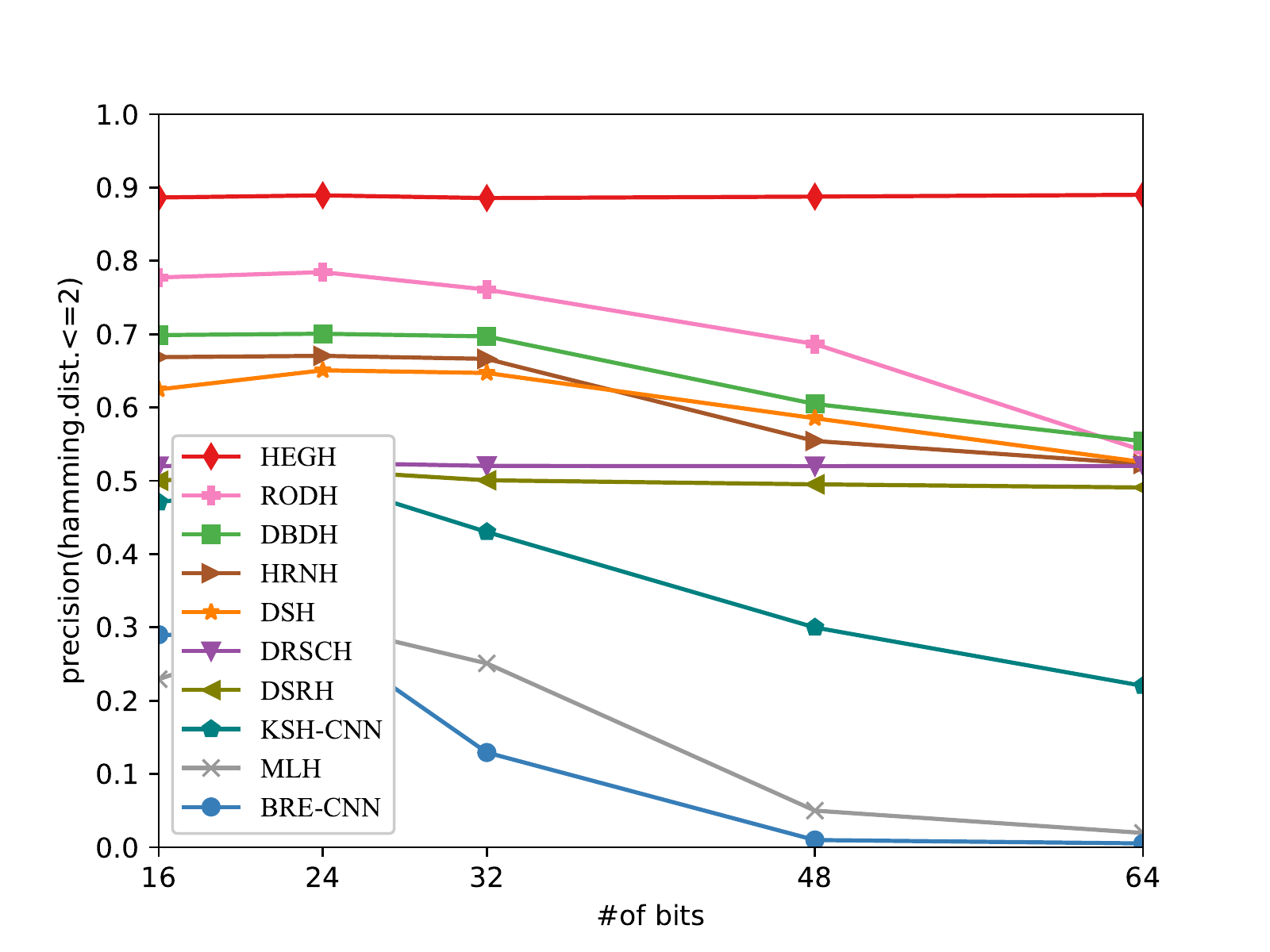}
    \caption{Precision curves within HAM2}\label{fig.10}
\end{figure}

\subsubsection{The T-SNE visualization}

Fig. \ref{fig.10} shows the T-SNE \cite{bib49} visualization of the hash codes generated by our HEGH, DBDH and DHN on CIFAR-10, respectively. As can be seen from the figures, the hash codes generated by DHN have obvious class clusters, but many examples are still scattered around or outside the clusters. These scattered examples may contain hard examples. In the T-SNE of DBDH. We find redundant clusters. This indicates that the semantics contained in the hash codes of some examples are different from those of most examples of the same kind, so additional clusters appear. The hash code generated by our method is greatly improved compared with other methods. Similar images are more clustered in the hash code cluster generated by HEGH, and there are more distinct margins between dissimilar images. Because we increase the penalty of hard example and make the hard examples scattered outside move closer to the class center. In addition, our network optimizes the semantic understanding ability of the network and extracts more key features to improve the learning of hard examples, which makes the distribution of the examples better in the feature space.

\begin{figure}[H]
    \centering
        \subfigure[DHN]{
        \label{Fig.11.a}
        \includegraphics[width=0.3\textwidth]{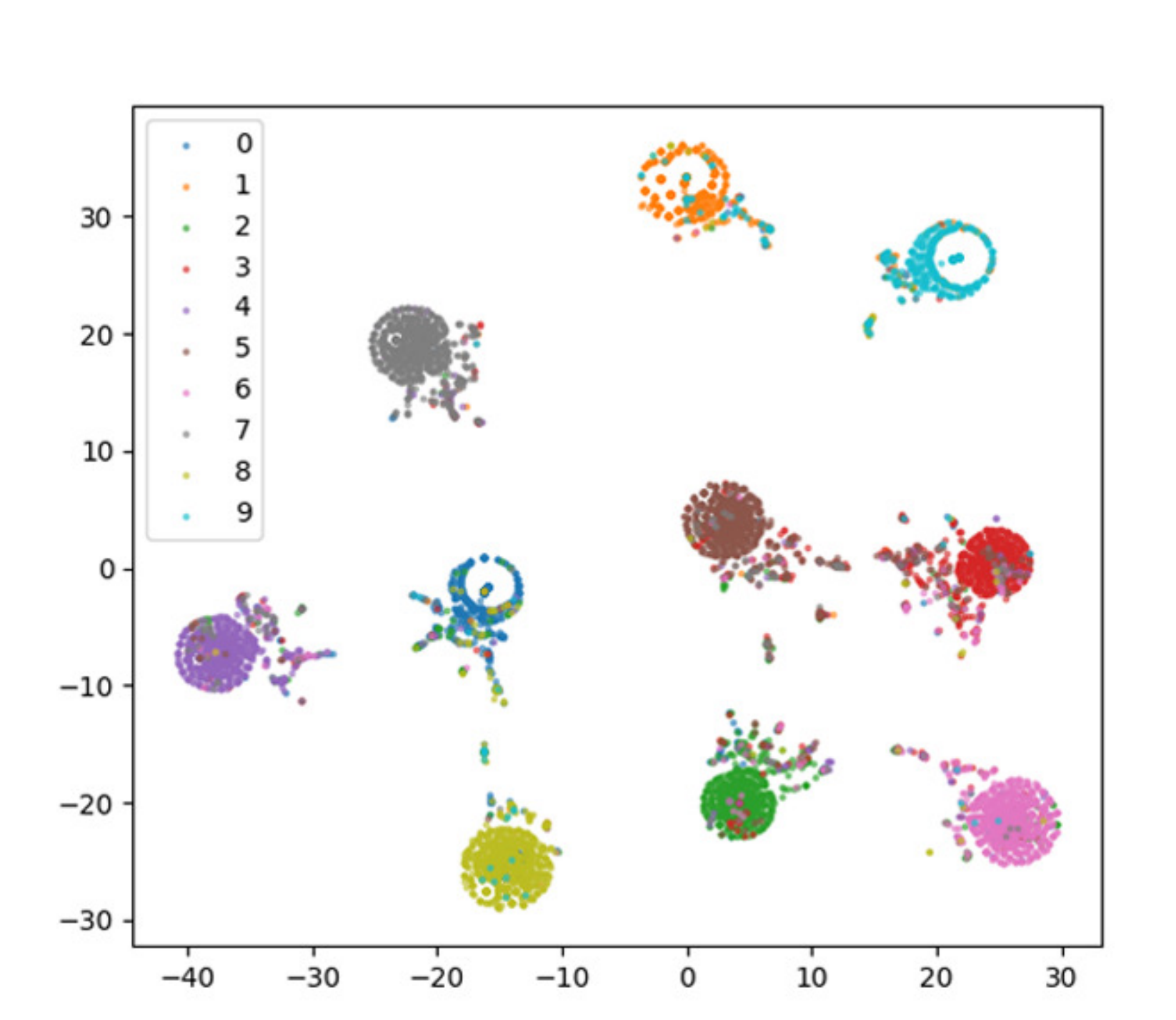}}
        \subfigure[DBDH]{
        \label{Fig.11.b}
        \includegraphics[width=0.3\textwidth]{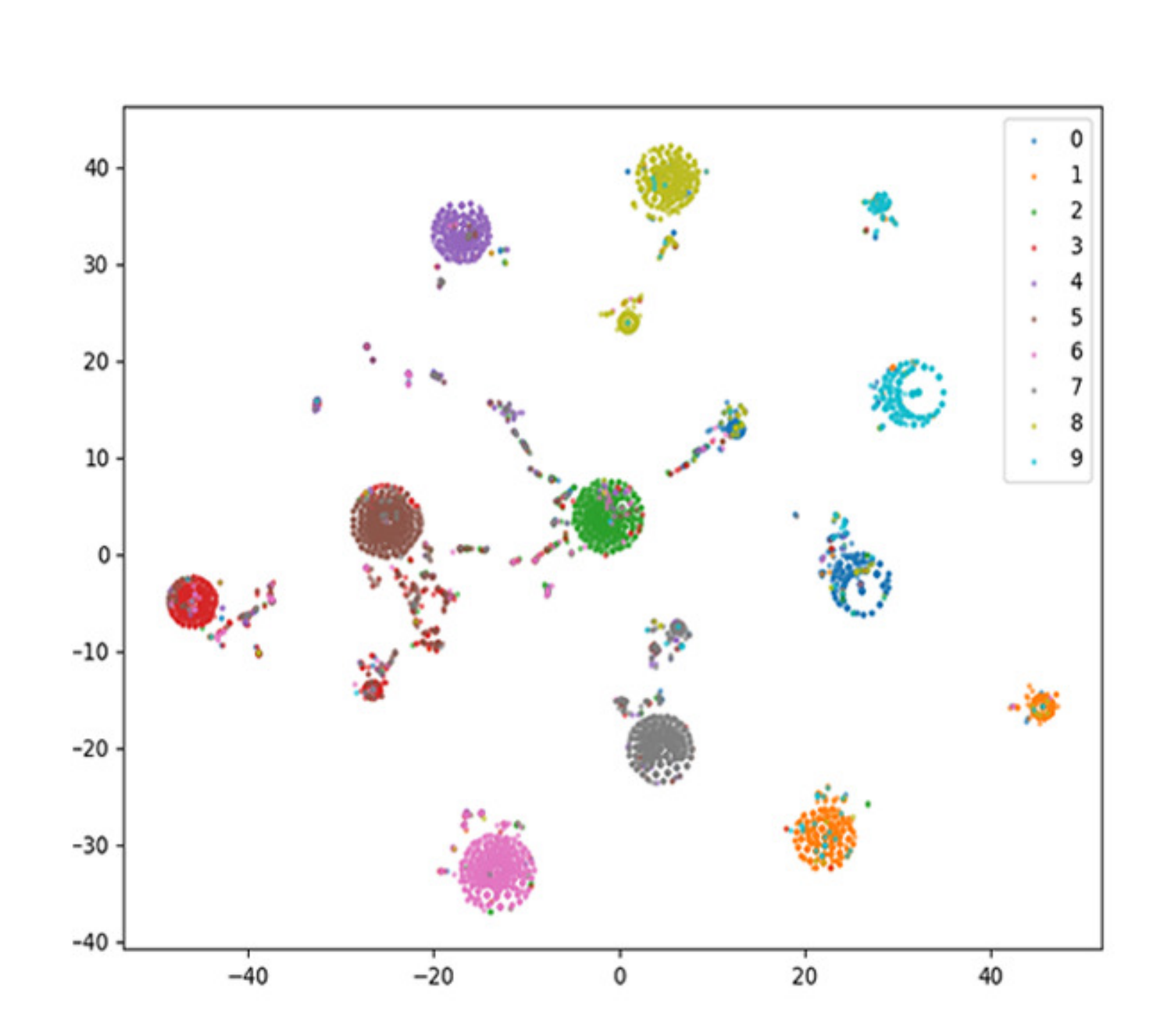}}
        \subfigure[HEGH]{
        \label{Fig.11.c}
        \includegraphics[width=0.3\textwidth]{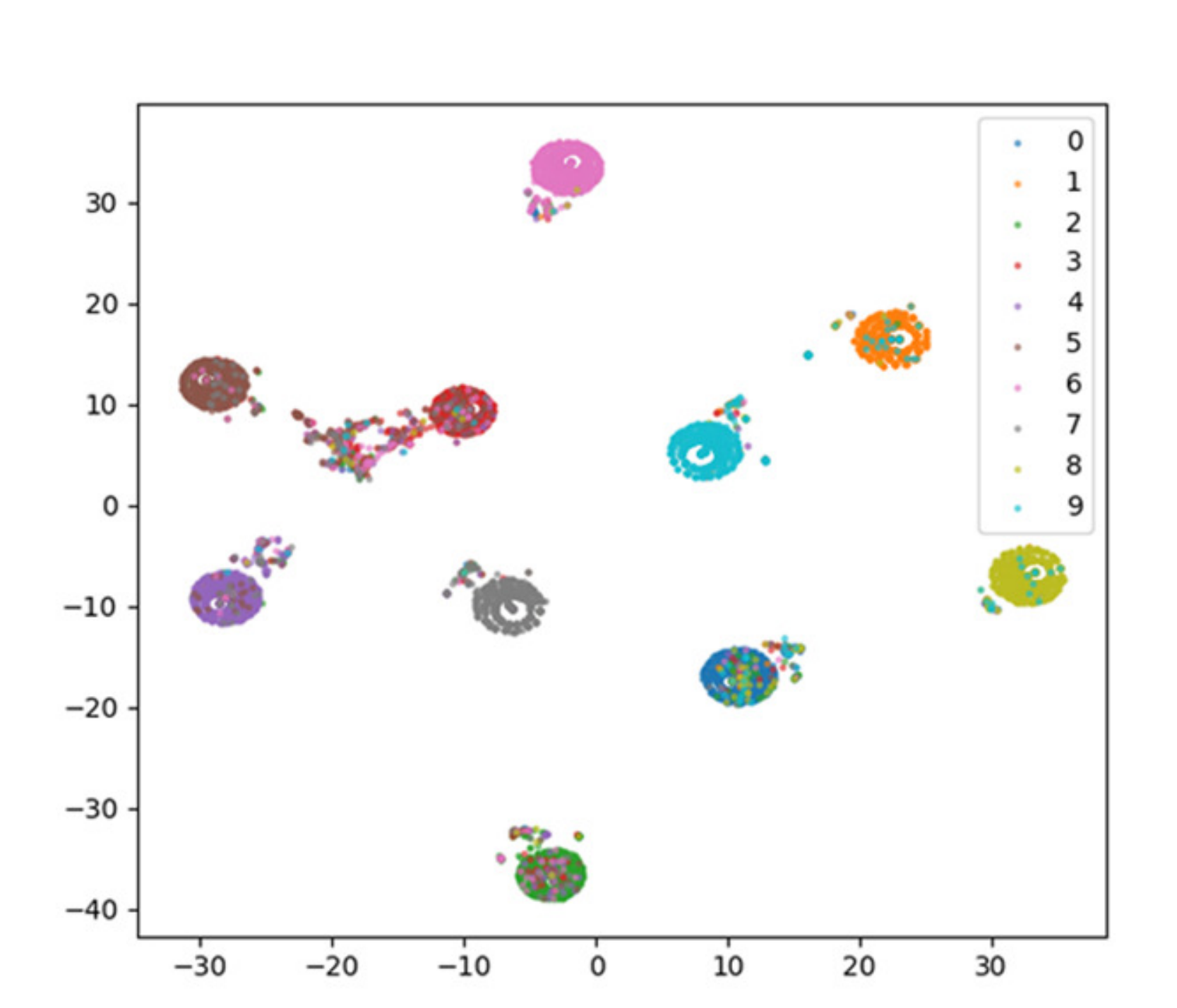}}
    \caption{The t-SNE visualizations of the hash codes on CIFAR-10}
    \label{Fig.11}
    \end{figure}

\section{Conclusion}\label{sec5}

Due to insufficient ability on capturing distinguished high-level features, deep hashing methods often incorrectly predict the similarity of hard examples. In this paper, we propose an end-to-end HEGH framework to solve the problem by improving the learning ability for hard examples. In our framework, CNN with CBAM module prevents key feature from being buried by a large number of redundant information so that the hard example can be effectively learning. Moreover, we make up for the quantity weakness of hard examples by modifying the penalty weights of examples to make hard examples being dominant in back propagation. Extensive experiments on two datasets demonstrate that the effectiveness of the proposed method outperforms the competing methods. In addition, it also exists some limitations on our work. For example, our method does not support the input of multi-size images. In the future work, we will further study the impact of hard examples on image retrieval and add new modules into our model to handle multi-size images so that it can efficiently accomplish the retrieval task in practical application.

\section*{Statements and Declarations}

\textbf{Conflict of interest} We, the authors of this paper, declare that we have no conflict of interest. The paper submitted is our own original workhas not been submitted to any other journal.
\bibliography{sn-bibliography}


\begin{thebibliography}{49}
\ifx \bisbn   \undefined \def \bisbn  #1{ISBN #1}\fi
\ifx \binits  \undefined \def \binits#1{#1}\fi
\ifx \bauthor  \undefined \def \bauthor#1{#1}\fi
\ifx \batitle  \undefined \def \batitle#1{#1}\fi
\ifx \bjtitle  \undefined \def \bjtitle#1{#1}\fi
\ifx \bvolume  \undefined \def \bvolume#1{\textbf{#1}}\fi
\ifx \byear  \undefined \def \byear#1{#1}\fi
\ifx \bissue  \undefined \def \bissue#1{#1}\fi
\ifx \bfpage  \undefined \def \bfpage#1{#1}\fi
\ifx \blpage  \undefined \def \blpage #1{#1}\fi
\ifx \burl  \undefined \def \burl#1{\textsf{#1}}\fi
\ifx \doiurl  \undefined \def \doiurl#1{\url{https://doi.org/#1}}\fi
\ifx \betal  \undefined \def \betal{\textit{et al.}}\fi
\ifx \binstitute  \undefined \def \binstitute#1{#1}\fi
\ifx \binstitutionaled  \undefined \def \binstitutionaled#1{#1}\fi
\ifx \bctitle  \undefined \def \bctitle#1{#1}\fi
\ifx \beditor  \undefined \def \beditor#1{#1}\fi
\ifx \bpublisher  \undefined \def \bpublisher#1{#1}\fi
\ifx \bbtitle  \undefined \def \bbtitle#1{#1}\fi
\ifx \bedition  \undefined \def \bedition#1{#1}\fi
\ifx \bseriesno  \undefined \def \bseriesno#1{#1}\fi
\ifx \blocation  \undefined \def \blocation#1{#1}\fi
\ifx \bsertitle  \undefined \def \bsertitle#1{#1}\fi
\ifx \bsnm \undefined \def \bsnm#1{#1}\fi
\ifx \bsuffix \undefined \def \bsuffix#1{#1}\fi
\ifx \bparticle \undefined \def \bparticle#1{#1}\fi
\ifx \barticle \undefined \def \barticle#1{#1}\fi
\bibcommenthead
\ifx \bconfdate \undefined \def \bconfdate #1{#1}\fi
\ifx \botherref \undefined \def \botherref #1{#1}\fi
\ifx \url \undefined \def \url#1{\textsf{#1}}\fi
\ifx \bchapter \undefined \def \bchapter#1{#1}\fi
\ifx \bbook \undefined \def \bbook#1{#1}\fi
\ifx \bcomment \undefined \def \bcomment#1{#1}\fi
\ifx \oauthor \undefined \def \oauthor#1{#1}\fi
\ifx \citeauthoryear \undefined \def \citeauthoryear#1{#1}\fi
\ifx \endbibitem  \undefined \def \endbibitem {}\fi
\ifx \bconflocation  \undefined \def \bconflocation#1{#1}\fi
\ifx \arxivurl  \undefined \def \arxivurl#1{\textsf{#1}}\fi
\csname PreBibitemsHook\endcsname

\bibitem{bib1}
\begin{botherref}
\oauthor{\bsnm{Ji}, \binits{J.}},
\oauthor{\bsnm{Li}, \binits{J.}},
\oauthor{\bsnm{Yan}, \binits{S.}},
\oauthor{\bsnm{Zhang}, \binits{B.}},
\oauthor{\bsnm{Tian}, \binits{Q.}}:
Super-bit locality-sensitive hashing,
108--116
(2012)
\end{botherref}
\endbibitem

\bibitem{bib2}
\begin{barticle}
\bauthor{\bsnm{Kulis}, \binits{B.}},
\bauthor{\bsnm{Grauman}, \binits{K.}}:
\batitle{Kernelized locality-sensitive hashing}.
\bjtitle{IEEE Transactions on Pattern Analysis and Machine Intelligence}
\bvolume{34}(\bissue{6}),
\bfpage{1092}--\blpage{1104}
(\byear{2011})
\end{barticle}
\endbibitem

\bibitem{bib3}
\begin{botherref}
\oauthor{\bsnm{Weiss}, \binits{Y.}},
\oauthor{\bsnm{Torralba}, \binits{A.}},
\oauthor{\bsnm{Fergus}, \binits{R.}}, et al.:
Spectral hashing.
\textbf{1}(2),
4
(2008).
Citeseer
\end{botherref}
\endbibitem

\bibitem{bib4}
\begin{barticle}
\bauthor{\bsnm{Zhu}, \binits{X.}},
\bauthor{\bsnm{Li}, \binits{X.}},
\bauthor{\bsnm{Zhang}, \binits{S.}},
\bauthor{\bsnm{Xu}, \binits{Z.}},
\bauthor{\bsnm{Yu}, \binits{L.}},
\bauthor{\bsnm{Wang}, \binits{C.}}:
\batitle{Graph pca hashing for similarity search}.
\bjtitle{IEEE Transactions on Multimedia}
\bvolume{19}(\bissue{9}),
\bfpage{2033}--\blpage{2044}
(\byear{2017})
\end{barticle}
\endbibitem

\bibitem{bib5}
\begin{botherref}
\oauthor{\bsnm{Wang}, \binits{Q.}},
\oauthor{\bsnm{Si}, \binits{L.}},
\oauthor{\bsnm{Zhang}, \binits{D.}}:
Learning to hash with partial tags: Exploring correlation between tags and
  hashing bits for large scale image retrieval,
378--392
(2014).
Springer
\end{botherref}
\endbibitem

\bibitem{bib6}
\begin{barticle}
\bauthor{\bsnm{Zhang}, \binits{R.}},
\bauthor{\bsnm{Lin}, \binits{L.}},
\bauthor{\bsnm{Zhang}, \binits{R.}},
\bauthor{\bsnm{Zuo}, \binits{W.}},
\bauthor{\bsnm{Zhang}, \binits{L.}}:
\batitle{Bit-scalable deep hashing with regularized similarity learning for
  image retrieval and person re-identification}.
\bjtitle{IEEE Transactions on Image Processing}
\bvolume{24}(\bissue{12}),
\bfpage{4766}--\blpage{4779}
(\byear{2015})
\end{barticle}
\endbibitem

\bibitem{bib7}
\begin{bchapter}
\bauthor{\bsnm{Lin}, \binits{K.}},
\bauthor{\bsnm{Yang}, \binits{H.-F.}},
\bauthor{\bsnm{Hsiao}, \binits{J.-H.}},
\bauthor{\bsnm{Chen}, \binits{C.-S.}}:
\bctitle{Deep learning of binary hash codes for fast image retrieval}.
In: \bbtitle{Proceedings of the IEEE Conference on Computer Vision and Pattern
  Recognition Workshops},
pp. \bfpage{27}--\blpage{35}
(\byear{2015})
\end{bchapter}
\endbibitem

\bibitem{bib8}
\begin{bchapter}
\bauthor{\bsnm{Lai}, \binits{H.}},
\bauthor{\bsnm{Pan}, \binits{Y.}},
\bauthor{\bsnm{Liu}, \binits{Y.}},
\bauthor{\bsnm{Yan}, \binits{S.}}:
\bctitle{Simultaneous feature learning and hash coding with deep neural
  networks}.
In: \bbtitle{Proceedings of the IEEE Conference on Computer Vision and Pattern
  Recognition},
pp. \bfpage{3270}--\blpage{3278}
(\byear{2015})
\end{bchapter}
\endbibitem

\bibitem{bib9}
\begin{bchapter}
\bauthor{\bsnm{Cao}, \binits{Y.}},
\bauthor{\bsnm{Long}, \binits{M.}},
\bauthor{\bsnm{Liu}, \binits{B.}},
\bauthor{\bsnm{Wang}, \binits{J.}}:
\bctitle{Deep cauchy hashing for hamming space retrieval}.
In: \bbtitle{Proceedings of the IEEE Conference on Computer Vision and Pattern
  Recognition},
pp. \bfpage{1229}--\blpage{1237}
(\byear{2018})
\end{bchapter}
\endbibitem

\bibitem{bib10}
\begin{bchapter}
\bauthor{\bsnm{Yan}, \binits{C.}},
\bauthor{\bsnm{Pang}, \binits{G.}},
\bauthor{\bsnm{Bai}, \binits{X.}},
\bauthor{\bsnm{Shen}, \binits{C.}},
\bauthor{\bsnm{Zhou}, \binits{J.}},
\bauthor{\bsnm{Hancock}, \binits{E.}}:
\bctitle{Deep hashing by discriminating hard examples}.
In: \bbtitle{Proceedings of the 27th ACM International Conference on
  Multimedia},
pp. \bfpage{1535}--\blpage{1542}
(\byear{2019})
\end{bchapter}
\endbibitem

\bibitem{bib11}
\begin{bchapter}
\bauthor{\bsnm{Weiss}, \binits{Y.}},
\bauthor{\bsnm{Fergus}, \binits{R.}},
\bauthor{\bsnm{Torralba}, \binits{A.}}:
\bctitle{Multidimensional spectral hashing}.
In: \bbtitle{European Conference on Computer Vision},
pp. \bfpage{340}--\blpage{353}
(\byear{2012}).
\bcomment{Springer}
\end{bchapter}
\endbibitem

\bibitem{bib12}
\begin{bchapter}
\bauthor{\bsnm{He}, \binits{K.}},
\bauthor{\bsnm{Wen}, \binits{F.}},
\bauthor{\bsnm{Sun}, \binits{J.}}:
\bctitle{K-means hashing: An affinity-preserving quantization method for
  learning binary compact codes}.
In: \bbtitle{Proceedings of the IEEE Conference on Computer Vision and Pattern
  Recognition},
pp. \bfpage{2938}--\blpage{2945}
(\byear{2013})
\end{bchapter}
\endbibitem

\bibitem{bib13}
\begin{barticle}
\bauthor{\bsnm{Shen}, \binits{F.}},
\bauthor{\bsnm{Xu}, \binits{Y.}},
\bauthor{\bsnm{Liu}, \binits{L.}},
\bauthor{\bsnm{Yang}, \binits{Y.}},
\bauthor{\bsnm{Huang}, \binits{Z.}},
\bauthor{\bsnm{Shen}, \binits{H.T.}}:
\batitle{Unsupervised deep hashing with similarity-adaptive and discrete
  optimization}.
\bjtitle{IEEE transactions on pattern analysis and machine intelligence}
\bvolume{40}(\bissue{12}),
\bfpage{3034}--\blpage{3044}
(\byear{2018})
\end{barticle}
\endbibitem

\bibitem{bib14}
\begin{botherref}
\oauthor{\bsnm{Yang}, \binits{H.-F.}},
\oauthor{\bsnm{Lin}, \binits{K.}},
\oauthor{\bsnm{Chen}, \binits{C.-S.}}:
Supervised learning of semantics-preserving hashing via deep neural networks
  for large-scale image search.
arXiv preprint arXiv:1507.00101
\textbf{2}
(2015)
\end{botherref}
\endbibitem

\bibitem{bib15}
\begin{bchapter}
\bauthor{\bsnm{Zhao}, \binits{F.}},
\bauthor{\bsnm{Huang}, \binits{Y.}},
\bauthor{\bsnm{Wang}, \binits{L.}},
\bauthor{\bsnm{Tan}, \binits{T.}}:
\bctitle{Deep semantic ranking based hashing for multi-label image retrieval}.
In: \bbtitle{Proceedings of the IEEE Conference on Computer Vision and Pattern
  Recognition},
pp. \bfpage{1556}--\blpage{1564}
(\byear{2015})
\end{bchapter}
\endbibitem

\bibitem{bib16}
\begin{bchapter}
\bauthor{\bsnm{Su}, \binits{S.}},
\bauthor{\bsnm{Zhang}, \binits{C.}},
\bauthor{\bsnm{Han}, \binits{K.}},
\bauthor{\bsnm{Tian}, \binits{Y.}}:
\bctitle{Greedy hash: Towards fast optimization for accurate hash coding in
  cnn}.
In: \bbtitle{Proceedings of the 32nd International Conference on Neural
  Information Processing Systems},
pp. \bfpage{806}--\blpage{815}
(\byear{2018})
\end{bchapter}
\endbibitem

\bibitem{bib17}
\begin{botherref}
\oauthor{\bsnm{Li}, \binits{Q.}},
\oauthor{\bsnm{Sun}, \binits{Z.}},
\oauthor{\bsnm{He}, \binits{R.}},
\oauthor{\bsnm{Tan}, \binits{T.}}:
Deep supervised discrete hashing.
arXiv preprint arXiv:1705.10999
(2017)
\end{botherref}
\endbibitem

\bibitem{bib18}
\begin{bchapter}
\bauthor{\bsnm{Xia}, \binits{R.}},
\bauthor{\bsnm{Pan}, \binits{Y.}},
\bauthor{\bsnm{Lai}, \binits{H.}},
\bauthor{\bsnm{Liu}, \binits{C.}},
\bauthor{\bsnm{Yan}, \binits{S.}}:
\bctitle{Supervised hashing for image retrieval via image representation
  learning}.
In: \bbtitle{Twenty-eighth AAAI Conference on Artificial Intelligence}
(\byear{2014})
\end{bchapter}
\endbibitem

\bibitem{bib19}
\begin{bchapter}
\bauthor{\bsnm{Erin~Liong}, \binits{V.}},
\bauthor{\bsnm{Lu}, \binits{J.}},
\bauthor{\bsnm{Wang}, \binits{G.}},
\bauthor{\bsnm{Moulin}, \binits{P.}},
\bauthor{\bsnm{Zhou}, \binits{J.}}:
\bctitle{Deep hashing for compact binary codes learning}.
In: \bbtitle{Proceedings of the IEEE Conference on Computer Vision and Pattern
  Recognition},
pp. \bfpage{2475}--\blpage{2483}
(\byear{2015})
\end{bchapter}
\endbibitem

\bibitem{bib20}
\begin{bchapter}
\bauthor{\bsnm{Li}, \binits{W.-J.}},
\bauthor{\bsnm{Wang}, \binits{S.}},
\bauthor{\bsnm{Kang}, \binits{W.-C.}}:
\bctitle{Feature learning based deep supervised hashing with pairwise labels}.
In: \bbtitle{Proceedings of the Twenty-Fifth International Joint Conference on
  Artificial Intelligence},
pp. \bfpage{1711}--\blpage{1717}
(\byear{2016})
\end{bchapter}
\endbibitem

\bibitem{bib21}
\begin{bchapter}
\bauthor{\bsnm{Cao}, \binits{Z.}},
\bauthor{\bsnm{Long}, \binits{M.}},
\bauthor{\bsnm{Wang}, \binits{J.}},
\bauthor{\bsnm{Philip}, \binits{S.Y.}}:
\bctitle{Hashnet: Deep learning to hash by continuation}.
In: \bbtitle{2017 IEEE International Conference on Computer Vision (ICCV)},
pp. \bfpage{5609}--\blpage{5618}
(\byear{2017}).
\bcomment{IEEE Computer Society}
\end{bchapter}
\endbibitem

\bibitem{bib22}
\begin{bchapter}
\bauthor{\bsnm{Shrivastava}, \binits{A.}},
\bauthor{\bsnm{Gupta}, \binits{A.}},
\bauthor{\bsnm{Girshick}, \binits{R.}}:
\bctitle{Training region-based object detectors with online hard example
  mining}.
In: \bbtitle{2016 IEEE Conference on Computer Vision and Pattern Recognition
  (CVPR)},
pp. \bfpage{761}--\blpage{769}
(\byear{2016}).
\bcomment{IEEE}
\end{bchapter}
\endbibitem

\bibitem{bib23}
\begin{bchapter}
\bauthor{\bsnm{Li}, \binits{M.}},
\bauthor{\bsnm{Zhang}, \binits{Z.}},
\bauthor{\bsnm{Yu}, \binits{H.}},
\bauthor{\bsnm{Chen}, \binits{X.}},
\bauthor{\bsnm{Li}, \binits{D.}}:
\bctitle{S-ohem: stratified online hard example mining for object detection}.
In: \bbtitle{CCF Chinese Conference on Computer Vision},
pp. \bfpage{166}--\blpage{177}
(\byear{2017}).
\bcomment{Springer}
\end{bchapter}
\endbibitem

\bibitem{bib24}
\begin{barticle}
\bauthor{\bsnm{Lin}, \binits{C.-T.}},
\bauthor{\bsnm{Chen}, \binits{S.-P.}},
\bauthor{\bsnm{Santoso}, \binits{P.S.}},
\bauthor{\bsnm{Lin}, \binits{H.-J.}},
\bauthor{\bsnm{Lai}, \binits{S.-H.}}:
\batitle{Real-time single-stage vehicle detector optimized by multi-stage
  image-based online hard example mining}.
\bjtitle{IEEE Transactions on Vehicular Technology}
\bvolume{69}(\bissue{2}),
\bfpage{1505}--\blpage{1518}
(\byear{2019})
\end{barticle}
\endbibitem

\bibitem{bib25}
\begin{barticle}
\bauthor{\bsnm{Cai}, \binits{B.}},
\bauthor{\bsnm{Jiang}, \binits{Z.}},
\bauthor{\bsnm{Zhang}, \binits{H.}},
\bauthor{\bsnm{Zhao}, \binits{D.}},
\bauthor{\bsnm{Yao}, \binits{Y.}}:
\batitle{Airport detection using end-to-end convolutional neural network with
  hard example mining}.
\bjtitle{Remote Sensing}
\bvolume{9}(\bissue{11}),
\bfpage{1198}
(\byear{2017})
\end{barticle}
\endbibitem

\bibitem{bib26}
\begin{barticle}
\bauthor{\bsnm{Chu}, \binits{J.}},
\bauthor{\bsnm{Guo}, \binits{Z.}},
\bauthor{\bsnm{Leng}, \binits{L.}}:
\batitle{Object detection based on multi-layer convolution feature fusion and
  online hard example mining}.
\bjtitle{IEEE Access}
\bvolume{6},
\bfpage{19959}--\blpage{19967}
(\byear{2018})
\end{barticle}
\endbibitem

\bibitem{bib27}
\begin{botherref}
\oauthor{\bsnm{Xiao}, \binits{Q.}},
\oauthor{\bsnm{Luo}, \binits{H.}},
\oauthor{\bsnm{Zhang}, \binits{C.}}:
Margin sample mining loss: A deep learning based method for person
  re-identification.
arXiv preprint arXiv:1710.00478
(2017)
\end{botherref}
\endbibitem

\bibitem{bib28}
\begin{bchapter}
\bauthor{\bsnm{Lin}, \binits{T.-Y.}},
\bauthor{\bsnm{Goyal}, \binits{P.}},
\bauthor{\bsnm{Girshick}, \binits{R.}},
\bauthor{\bsnm{He}, \binits{K.}},
\bauthor{\bsnm{Doll{\'a}r}, \binits{P.}}:
\bctitle{Focal loss for dense object detection}.
In: \bbtitle{Proceedings of the IEEE International Conference on Computer
  Vision},
pp. \bfpage{2980}--\blpage{2988}
(\byear{2017})
\end{bchapter}
\endbibitem

\bibitem{bib29}
\begin{barticle}
\bauthor{\bsnm{Zhang}, \binits{F.}},
\bauthor{\bsnm{Du}, \binits{B.}},
\bauthor{\bsnm{Zhang}, \binits{L.}},
\bauthor{\bsnm{Xu}, \binits{M.}}:
\batitle{Weakly supervised learning based on coupled convolutional neural
  networks for aircraft detection}.
\bjtitle{IEEE Transactions on Geoscience and Remote Sensing}
\bvolume{54}(\bissue{9}),
\bfpage{5553}--\blpage{5563}
(\byear{2016})
\end{barticle}
\endbibitem

\bibitem{bib30}
\begin{botherref}
\oauthor{\bsnm{Hermans}, \binits{A.}},
\oauthor{\bsnm{Beyer}, \binits{L.}},
\oauthor{\bsnm{Leibe}, \binits{B.}}:
In defense of the triplet loss for person re-identification.
arXiv preprint arXiv:1703.07737
(2017)
\end{botherref}
\endbibitem

\bibitem{bib31}
\begin{barticle}
\bauthor{\bsnm{Chen}, \binits{Y.}},
\bauthor{\bsnm{Lu}, \binits{X.}}:
\batitle{Deep discrete hashing with pairwise correlation learning}.
\bjtitle{Neurocomputing}
\bvolume{385},
\bfpage{111}--\blpage{121}
(\byear{2020})
\end{barticle}
\endbibitem

\bibitem{bib32}
\begin{bchapter}
\bauthor{\bsnm{Shen}, \binits{F.}},
\bauthor{\bsnm{Shen}, \binits{C.}},
\bauthor{\bsnm{Liu}, \binits{W.}},
\bauthor{\bsnm{Tao~Shen}, \binits{H.}}:
\bctitle{Supervised discrete hashing}.
In: \bbtitle{Proceedings of the IEEE Conference on Computer Vision and Pattern
  Recognition},
pp. \bfpage{37}--\blpage{45}
(\byear{2015})
\end{bchapter}
\endbibitem

\bibitem{bib33}
\begin{barticle}
\bauthor{\bsnm{Krizhevsky}, \binits{A.}},
\bauthor{\bsnm{Sutskever}, \binits{I.}},
\bauthor{\bsnm{Hinton}, \binits{G.E.}}:
\batitle{Imagenet classification with deep convolutional neural networks}.
\bjtitle{Advances in neural information processing systems}
\bvolume{25},
\bfpage{1097}--\blpage{1105}
(\byear{2012})
\end{barticle}
\endbibitem

\bibitem{bib34}
\begin{botherref}
\oauthor{\bsnm{Simonyan}, \binits{K.}},
\oauthor{\bsnm{Zisserman}, \binits{A.}}:
Very deep convolutional networks for large-scale image recognition.
arXiv preprint arXiv:1409.1556
(2014)
\end{botherref}
\endbibitem

\bibitem{bib35}
\begin{bchapter}
\bauthor{\bsnm{Liu}, \binits{H.}},
\bauthor{\bsnm{Wang}, \binits{R.}},
\bauthor{\bsnm{Shan}, \binits{S.}},
\bauthor{\bsnm{Chen}, \binits{X.}}:
\bctitle{Deep supervised hashing for fast image retrieval}.
In: \bbtitle{Proceedings of the IEEE Conference on Computer Vision and Pattern
  Recognition},
pp. \bfpage{2064}--\blpage{2072}
(\byear{2016})
\end{bchapter}
\endbibitem

\bibitem{bib36}
\begin{bchapter}
\bauthor{\bsnm{Woo}, \binits{S.}},
\bauthor{\bsnm{Park}, \binits{J.}},
\bauthor{\bsnm{Lee}, \binits{J.-Y.}},
\bauthor{\bsnm{Kweon}, \binits{I.S.}}:
\bctitle{Cbam: Convolutional block attention module}.
In: \bbtitle{Proceedings of the European Conference on Computer Vision (ECCV)},
pp. \bfpage{3}--\blpage{19}
(\byear{2018})
\end{bchapter}
\endbibitem

\bibitem{bib37}
\begin{barticle}
\bauthor{\bsnm{Jin}, \binits{Z.}},
\bauthor{\bsnm{Li}, \binits{C.}},
\bauthor{\bsnm{Lin}, \binits{Y.}},
\bauthor{\bsnm{Cai}, \binits{D.}}:
\batitle{Density sensitive hashing}.
\bjtitle{IEEE transactions on cybernetics}
\bvolume{44}(\bissue{8}),
\bfpage{1362}--\blpage{1371}
(\byear{2013})
\end{barticle}
\endbibitem

\bibitem{bib38}
\begin{barticle}
\bauthor{\bsnm{Zhang}, \binits{Z.}},
\bauthor{\bsnm{Lai}, \binits{Z.}},
\bauthor{\bsnm{Huang}, \binits{Z.}},
\bauthor{\bsnm{Wong}, \binits{W.K.}},
\bauthor{\bsnm{Xie}, \binits{G.-S.}},
\bauthor{\bsnm{Liu}, \binits{L.}},
\bauthor{\bsnm{Shao}, \binits{L.}}:
\batitle{Scalable supervised asymmetric hashing with semantic and latent factor
  embedding}.
\bjtitle{IEEE Transactions on Image Processing}
\bvolume{28}(\bissue{10}),
\bfpage{4803}--\blpage{4818}
(\byear{2019})
\end{barticle}
\endbibitem

\bibitem{bib39}
\begin{barticle}
\bauthor{\bsnm{Zheng}, \binits{X.}},
\bauthor{\bsnm{Zhang}, \binits{Y.}},
\bauthor{\bsnm{Lu}, \binits{X.}}:
\batitle{Deep balanced discrete hashing for image retrieval}.
\bjtitle{Neurocomputing}
\bvolume{403},
\bfpage{224}--\blpage{236}
(\byear{2020})
\end{barticle}
\endbibitem

\bibitem{bib40}
\begin{barticle}
\bauthor{\bsnm{Li}, \binits{Q.}},
\bauthor{\bsnm{Sun}, \binits{Z.}},
\bauthor{\bsnm{He}, \binits{R.}},
\bauthor{\bsnm{Tan}, \binits{T.}}:
\batitle{A general framework for deep supervised discrete hashing}.
\bjtitle{International Journal of Computer Vision}
\bvolume{128}(\bissue{8}),
\bfpage{2204}--\blpage{2222}
(\byear{2020})
\end{barticle}
\endbibitem

\bibitem{bib41}
\begin{barticle}
\bauthor{\bsnm{Lu}, \binits{X.}},
\bauthor{\bsnm{Chen}, \binits{Y.}},
\bauthor{\bsnm{Li}, \binits{X.}}:
\batitle{Discrete deep hashing with ranking optimization for image retrieval}.
\bjtitle{IEEE transactions on neural networks and learning systems}
\bvolume{31}(\bissue{6}),
\bfpage{2052}--\blpage{2063}
(\byear{2019})
\end{barticle}
\endbibitem

\bibitem{bib42}
\begin{botherref}
\oauthor{\bsnm{Zhang}, \binits{B.}},
\oauthor{\bsnm{Qian}, \binits{J.}},
\oauthor{\bsnm{Xie}, \binits{X.}},
\oauthor{\bsnm{Xin}, \binits{Y.}},
\oauthor{\bsnm{Dong}, \binits{Y.}}:
Capsnet-based supervised hashing.
Applied Intelligence,
1--15
(2021)
\end{botherref}
\endbibitem

\bibitem{bib43}
\begin{botherref}
\oauthor{\bsnm{Li}, \binits{Y.}},
\oauthor{\bsnm{Pei}, \binits{W.}},
\oauthor{\bparticle{van} \bsnm{Gemert}, \binits{J.}}, et al.:
Push for quantization: Deep fisher hashing.
arXiv preprint arXiv:1909.00206
(2019)
\end{botherref}
\endbibitem

\bibitem{bib44}
\begin{barticle}
\bauthor{\bsnm{Lu}, \binits{X.}},
\bauthor{\bsnm{Chen}, \binits{Y.}},
\bauthor{\bsnm{Li}, \binits{X.}}:
\batitle{Hierarchical recurrent neural hashing for image retrieval with
  hierarchical convolutional features}.
\bjtitle{IEEE transactions on image processing}
\bvolume{27}(\bissue{1}),
\bfpage{106}--\blpage{120}
(\byear{2017})
\end{barticle}
\endbibitem

\bibitem{bib45}
\begin{bchapter}
\bauthor{\bsnm{Liu}, \binits{W.}},
\bauthor{\bsnm{Wang}, \binits{J.}},
\bauthor{\bsnm{Ji}, \binits{R.}},
\bauthor{\bsnm{Jiang}, \binits{Y.-G.}},
\bauthor{\bsnm{Chang}, \binits{S.-F.}}:
\bctitle{Supervised hashing with kernels}.
In: \bbtitle{2012 IEEE Conference on Computer Vision and Pattern Recognition},
pp. \bfpage{2074}--\blpage{2081}
(\byear{2012}).
\bcomment{IEEE}
\end{bchapter}
\endbibitem

\bibitem{bib46}
\begin{bchapter}
\bauthor{\bsnm{Norouzi}, \binits{M.}},
\bauthor{\bsnm{Fleet}, \binits{D.J.}}:
\bctitle{Minimal loss hashing for compact binary codes}.
In: \bbtitle{ICML}
(\byear{2011})
\end{bchapter}
\endbibitem

\bibitem{bib47}
\begin{bchapter}
\bauthor{\bsnm{Kulis}, \binits{B.}},
\bauthor{\bsnm{Darrell}, \binits{T.}}:
\bctitle{Learning to hash with binary reconstructive embeddings.}
In: \bbtitle{NIPS},
vol. \bseriesno{22},
pp. \bfpage{1042}--\blpage{1050}
(\byear{2009}).
\bcomment{Citeseer}
\end{bchapter}
\endbibitem

\bibitem{bib48}
\begin{barticle}
\bauthor{\bsnm{Yang}, \binits{J.}},
\bauthor{\bsnm{Zhang}, \binits{Y.}},
\bauthor{\bsnm{Feng}, \binits{R.}},
\bauthor{\bsnm{Zhang}, \binits{T.}},
\bauthor{\bsnm{Fan}, \binits{W.}}:
\batitle{Deep reinforcement hashing with redundancy elimination for effective
  image retrieval}.
\bjtitle{Pattern Recognition}
\bvolume{100},
\bfpage{107116}
(\byear{2020})
\end{barticle}
\endbibitem

\bibitem{bib49}
\begin{botherref}
\oauthor{\bparticle{Van~der} \bsnm{Maaten}, \binits{L.}},
\oauthor{\bsnm{Hinton}, \binits{G.}}:
Visualizing data using t-sne.
Journal of machine learning research
\textbf{9}(11)
(2008)
\end{botherref}
\endbibitem

\end{thebibliography}

\end{document}